\documentclass[times, review, 10pt]{elsarticle} 

\usepackage{amssymb}
\usepackage{amsmath}

\usepackage[numbers]{natbib}
\usepackage{booktabs} 
\usepackage{multirow}
\usepackage{graphicx}
\usepackage{array}
\usepackage{rotating}
\usepackage{colortbl, xcolor}
\usepackage[table]{xcolor}
\usepackage{hyperref}
\usepackage{subcaption}
\usepackage{tabularx}
\usepackage{cleveref}
\usepackage{xspace}
\usepackage{makecell}

\makeatletter
\DeclareRobustCommand\onedot{\futurelet\@let@token\@onedot}
\def\@onedot{\ifx\@let@token.\else.\null\fi\xspace}

\def\eg{\emph{e.g}\onedot,\xspace}

\def\ie{\emph{i.e}\onedot,\xspace}

\journal{Pattern Recognition}

\begin{document}

\begin{frontmatter}

\title{Decoupling Augmentation Bias in Prompt Learning for Vision-Language Models}                      
\author[1]{Gahyeon Kim\fnref{equal1}}
\author[1]{Sohee Kim\fnref{equal1}}
\author[1]{Seokju Lee\corref{cor1}}
\ead{slee@kentech.ac.kr}
\cortext[cor1]{Corresponding author.}
\fntext[equal1]{These authors contributed equally to this work.}

\affiliation[1]{
                organization={Korea Institute of Energy Technology (KENTECH)},
                country={Republic of Korea}
                }
                

                
\begin{abstract}
Recent advances in large-scale vision and language models have led to significant progress in zero-shot learning tasks. Methods such as CoOp and CoCoOp have shown that replacing handcrafted prompts with learnable vectors, known as prompt learning, can result in improved performance. However, these models often struggle to generalize to entirely unseen categories. While traditional zero-shot learning techniques benefit from various data augmentation strategies, prompt learning has primarily focused on text-based modifications, leaving the potential of image-based augmentation largely unexplored.
In this work, we explore how image-level augmentations, particularly those that introduce attribute-specific variations, can support and enhance prompt learning. Our analysis examines the interaction between these augmentations and soft prompt frameworks, revealing their potential to improve generalization. We also identify a limitation in existing methods, such as CoCoOp, which do not provide explicit guidance for learning prompts that focus on semantically meaningful visual features.
To address this, we propose Adding Attributes to Prompt Learning, AAPL, a novel method that introduces adversarial token embeddings to decouple superficial visual variations introduced by augmentation from class-relevant semantic representations. This decoupling enables the learned prompts to concentrate on visually discriminative features that align with the target categories. We conduct comprehensive experiments on eleven benchmark datasets, and AAPL consistently outperforms existing methods across few-shot, zero-shot, cross-dataset, and domain generalization settings. Our source code is publicly available at: 
\url{https://github.com/Gahyeonkim09/AAPL}
\end{abstract}








\end{frontmatter}

\section{Introduction}
\label{section1:introduction} 
\sloppy
Recent advances in large-scale vision-language models (VLMs), such as CLIP~\cite{radford2021learning}, have demonstrated strong image-text alignment via contrastive learning, yielding remarkable performance in zero-shot classification~\cite{zhang2022contrastive,singh2022flava,yuan2021florence}. Despite these successes, such models rely on fixed, hand-crafted prompts that are unstable, highly sensitive to subtle textual variations, and often demand substantial manual effort, thereby limiting their adaptability across diverse tasks and domains.

To address this limitation, CoOp~\cite{zhou2022learning} introduced soft prompt tuning, replacing static prompts with learnable vectors to enable efficient adaptation of frozen VLMs. Building on this idea, CoCoOp~\cite{zhou2022conditional} further enhanced adaptability by generating prompts conditioned on image features, improving class-specific performance. While these studies marked a significant shift toward prompt learning, most subsequent approaches have continued to depend primarily on textual cues or precomputed visual features, with limited exploration of integrating image augmentation into prompt optimization.

Extending the idea of prompt diversification, methods like PromDA~\cite{wang2022promda} and MixPro~\cite{li2025mixpro} incorporate text-based augmentation strategies, including sampling and mixup, to enhance few-shot robustness. Similarly, approaches like DUDE~\cite{nguyen2024dude} and CLAP~\cite{cai2024clap} integrate visual elements into prompt construction, aiming to better align visual and textual representations through structured tokens and contrastive learning objectives.

The limited use of augmentation, primarily as a preprocessing step or an auxiliary input, prevents its joint optimization with prompt representations and attribute-level features. As a result, existing systems often conflate class-relevant semantics with incidental attribute variations, \eg background, texture, and style, introducing biases that degrade generalization in few-shot and cross-domain scenarios. 
Most prior approaches do not explicitly model or regulate such attribute-level variations within the prompt learning process, leaving a gap in addressing fine-grained appearance changes while preserving semantic integrity. In methods like CoCoOp, the influence of image-derived conditional bias on the learnable prompt remains difficult to interpret or control, raising concerns about unintended biases in challenging generalization settings~\cite{ma2023understanding,khattak2023self,liu2023hierarchical}.

\begin{figure}[t]
  \centering
  \includegraphics[width=0.85\linewidth]{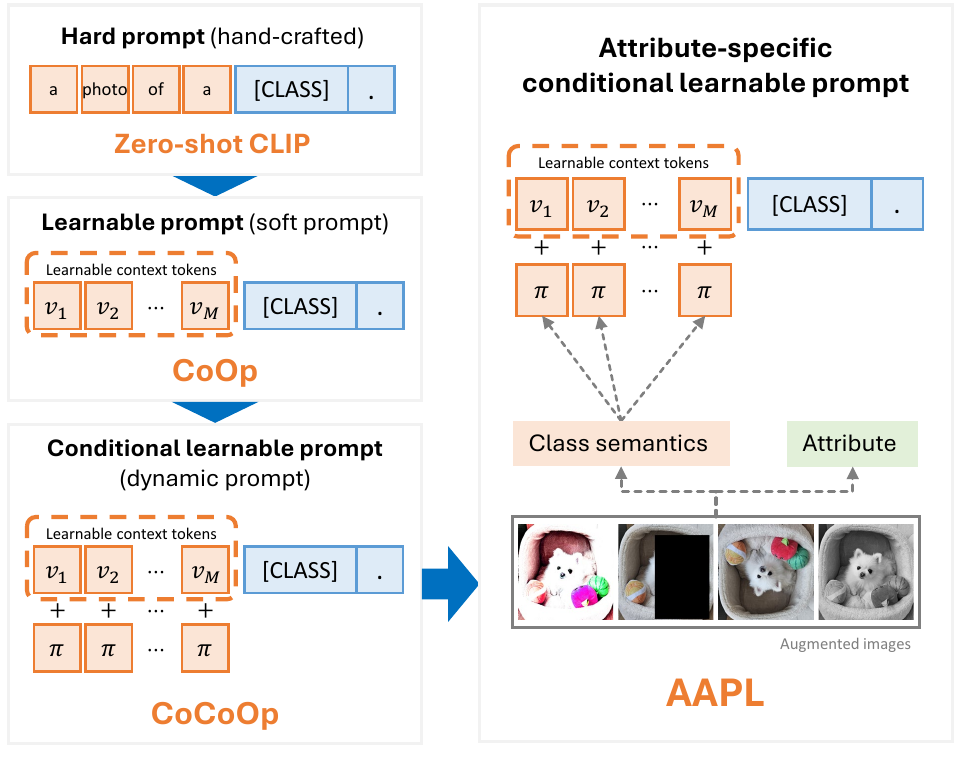}    
  \caption{\textbf{Comparison of prompt learning strategies in vision-language models.} 
  Zero-shot CLIP employs fixed, hand-crafted hard prompts, while CoOp replaces them with learnable soft prompts. CoCoOp further enhances prompt learning by introducing instance-specific biases through dynamic prompts. The proposed method, AAPL, proposes attribute-specific conditional learnable prompts that decompose image features into class semantics and attributes, injecting attribute-guided bias into the prompt. By leveraging attribute-specific information, AAPL improves adaptability to diverse contexts, leading to improved generalization and performance on unseen tasks.
  } 
  \label{fig:fig1}
\end{figure}

To address these limitations, we propose AAPL, ``Adding Attributes to Prompt Learning,'' a novel framework that systematically incorporates image augmentation as a visual prompt, illustrated in Fig.~\ref{fig:fig1}. Instead of passively conditioning prompts on raw image features, AAPL encodes attribute-specific variations derived from controlled image perturbations into the prompt space. This is achieved through an adversarial token embedding mechanism, which decouples low-level augmentation features from high-level semantic content, enabling the model to focus on meaningful attributes while suppressing overfitting to irrelevant visual noise.
Our method introduces the \textit{delta meta token}, a dedicated representation that captures attribute-induced variation. Using an adversarial triplet loss, we further enforce semantic consistency in the conditional bias across augmented views. As a result, AAPL enhances the model’s ability to generalize across attribute-rich domains, novel compositions, and unseen class distributions.

In summary, our contributions are as follows:
\begin{itemize}
\item We introduce \textbf{AAPL} (\emph{Adding Attributes to Prompt Learning}), which employs adversarial token embeddings to decouple low-level appearance variations caused by image augmentations from class-relevant semantics. This enables soft prompts to focus on discriminative and consistent visual semantics.

\item We propose a \emph{delta meta token} and an \emph{AdTriplet loss} to explicitly regulate the conditional bias of prompts. Through detailed augmentation profiling, we analyze and verify how these components promote semantic consistency while suppressing noise from attribute-level transformations.

\item Evaluated on 11 benchmark datasets, AAPL achieves competitive performance with strong baselines in zero-shot, few-shot, cross-dataset, and domain generalization tasks, showing comparable or better results.
\end{itemize}

\section{Related Works}
\label{section2:related_work}

\subsection{Vision-language models}  
\label{section2_subsec1:vlms} 
Vision-language models (VLMs) trained on image-text pairs have shown superior capabilities compared to image-only models, particularly in zero-shot transfer across diverse downstream classification tasks. Prominent models, such as CLIP~\cite{radford2021learning} and ALIGN~\cite{jia2021scaling}, leverage large-scale web data and employ self-supervised contrastive learning to align textual and visual representations. 
The contrastive loss in the embedding space brings matched image-text pairs closer together and pushes mismatched pairs apart, enabling strong generalization to unseen classes. 
In more general multi-modal settings, other deep learning methods have also shown how useful it is to combine different types of data and domain-specific visual features for robust classification and detection~\cite{sunil2023feature,sagari2024euri}.

Among these, CLIP~\cite{radford2021learning} stands out for being trained on 400 million image-text pairs, achieving remarkable zero-shot recognition performance without additional fine-tuning. Such large-scale pre-training enables VLMs to learn robust and transferable representations that generalize well across diverse domains. 
However, applying these broadly trained models to specialized tasks remains challenging, as real-world applications often require fine-grained adaptation beyond the original training distribution.

Our goal is to investigate efficient adaptation strategies for pre-trained VLMs, particularly in prompt learning. Beyond leveraging their inherent strengths, we aim to develop novel techniques that enhance adaptability to task-specific scenarios. By optimizing prompt-model interactions and refining embedding alignment, we seek to maximize their effectiveness across a broad range of real-world applications.

\subsection{Prompt Learning in Vision-Language Models}
\label{section2_subsec2:promptlearning in vlms}
Prompt learning, originally developed in natural language processing (NLP) to improve fine-tuning efficiency by replacing hand-crafted prompts with learnable embeddings, has been extended to vision-language models (VLMs) for adaptation to diverse tasks with minimal supervision.
An early influential method in prompt learning for VLMs is Context Optimization (CoOp)~\cite{zhou2022learning}, 
which replaces fixed textual templates with continuous context vectors optimized during training. 
While CoOp achieves strong performance in-domain, it often overfits in few-shot scenarios and struggles to generalize across domains.
To address these limitations, several extensions have been proposed.  
Distribution-aware approaches such as ProDA~\cite{lu2022prompt} and ProGrad~\cite{zhu2023prompt} improve robustness by modeling prompt distributions or selectively updating prompts whose gradients align with CLIP’s general knowledge.  
Knowledge-guided methods like KgCoOp~\cite{yao2023visual} regularize learning using class-level textual descriptions.  
Other works introduce structural or regularization-based improvements: FA~\cite{lu2025fa} adds a learnable forced prompt to complement a frozen original prompt for better in-/out-of-distribution performance, ProMetaR~\cite{park2024promptr} applies meta-regularization, CoPrompt~\cite{roy2024coprompt} enforces multi-view consistency, and ATPrompt~\cite{li2025advancing} uses attribute–category hybrid prompts for more discriminative textual features.

Beyond the text-only paradigm, image-conditioned prompt learning adapts prompts dynamically based on visual features.  
Conditional Context Optimization (CoCoOp)~\cite{zhou2022conditional} generates instance-specific prompt tokens from image features via a meta-network, improving generalization to unseen classes.  
More recently, A$^3$~\cite{wang2025a3} incorporates cross-modal adversarial feature alignment to mitigate the effect of unlearnable examples.

Prompt learning in the visual domain begins with Visual Prompt Tuning (VPT)~\cite{jia2022vpt}, which learns visual-only prompts by inserting tokens into the vision transformer. 
MaPLe~\cite{khattak2023maple} jointly optimizes text and visual prompts to improve cross-modal alignment. 
PromptKD~\cite{li2024promptkd} applies knowledge distillation to visual prompt learning, and MPL~\cite{huang2024modular} adapts hierarchical textual prompting to visual tokens. 
Recent methods include HiCroPL~\cite{zheng2025hierarchical} for hierarchical prompting, DiMPLe~\cite{rahman2025dimple} for disentangled representations.

Other lightweight adaptation methods, such as CLIP-Adapter~\cite{gao2024clip} and Tip-Adapter~\cite{zhang2021tip}, enable rapid domain adaptation with minimal parameter updates, complementing prompt learning approaches. Overall, research has shifted from architectural modifications toward designing more effective prompts that leverage both visual and textual signals, yielding improvements in few-shot learning, cross-domain generalization, and visual reasoning in real-world settings. Our method also adopts an image-conditioned prompt design similar to CoCoOp, enabling dynamic prompt generation from visual features for enhanced adaptability.

\begin{table}[t]
\centering
\footnotesize
\setlength{\tabcolsep}{4pt}
\renewcommand{\arraystretch}{1}
\begin{tabularx}{\linewidth}{p{0.42\linewidth} X}
\toprule
\textbf{Approach} & \textbf{Methods \textcolor{gray}{(Year)}} \\
\midrule
Text-only prompt learning \textcolor{gray}{(trainable textual prompts, without visual feature conditioning)} &
CoOp~\textcolor{gray}{('22)}, ProDA~\textcolor{gray}{('22)}, ProGrad~\textcolor{gray}{('23)}, KgCoOp~\textcolor{gray}{('23)}, DUDE~\textcolor{gray}{('24)}, FA~\textcolor{gray}{('25)}, ProMetaR~\textcolor{gray}{('24)}, CoPrompt~\textcolor{gray}{('24)}, ATPrompt~\textcolor{gray}{('25)} \\
\midrule
Image-conditioned text prompt learning \textcolor{gray}{(textual prompts conditioned on visual features)} &
CoCoOp~\textcolor{gray}{('22)}, A$^3$~\textcolor{gray}{('25)} \\
\midrule
Text-visual prompt learning \textcolor{gray}{(trainable textual \& visual prompts)} &
LoGoPrompt~\textcolor{gray}{('23)}, MaPLe~\textcolor{gray}{('23)}, PromptKD~\textcolor{gray}{('24)}, MPL~\textcolor{gray}{('24)}, HiCroPL~\textcolor{gray}{('25)}, DiMPLe~\textcolor{gray}{('25)}, AugPT~\textcolor{gray}{('25)} \\
\bottomrule
\end{tabularx}
\caption{\textbf{Survey of prompt learning methods} categorized by the type of trainable prompts and whether visual features are used for conditioning. 
Text-only methods update only textual prompts, image-conditioned text methods adapt textual prompts conditioned on visual features, 
and text-visual methods optimize both textual and visual prompts for better cross-modal alignment and robustness. }
\label{tab:related_work_table}
\end{table}

\subsection{Augmentation in Prompt Learning}
\label{section2_subsec3:augmentation in prompt learning}
Data augmentation has been explored to enhance prompt diversity and generalization, especially in low-resource and few-shot settings. PromDA~\cite{wang2022promda} creates diverse prompts through text-based sampling, while MixPro~\cite{li2025mixpro} augments prompts by mixing and recombining them at the sentence and token levels, yielding measurable gains. CLAP~\cite{cai2024clap} applies prompt-based textual augmentation to introduce stylistic variation, using contrastive learning to disentangle semantic content from superficial attributes such as style or texture.

Other approaches exploit visual information in different ways. 
LoGoPrompt~\cite{shi2023logoprompt} directly generates class-specific, text-rendered images to serve as visual prompts. 
DUDE~\cite{nguyen2024dude} aligns domain/class tokens with visual features to improve semantic consistency, and AugPT~\cite{li2025raw} incorporates self-supervised visual augmentation into prompt tuning, using a gating mechanism to filter noisy views and improve robustness without external data.

Despite these advances, few frameworks directly integrate image-level augmentation into prompt optimization, with most treating it as preprocessing or auxiliary input. This limits the modeling of fine-grained, attribute-level variations critical for zero-shot generalization. We address this by explicitly modeling attribute-specific variation in the prompt space, encoding structured cues from controlled visual perturbations to disentangle semantic identity from visual variability, thereby enabling better generalization in attribute-rich, fine-grained domains. 

An overview of the discussed works is shown in Table~\ref{tab:related_work_table}, which groups methods by trainable prompt modality and conditioning: text-only prompt learning optimizes textual prompts without visual feature conditioning, image-conditioned text prompt learning adapts textual prompts using visual features to improve generalization, and text-visual prompt learning optimizes textual and visual prompts for cross-modal alignment and robustness.

\subsection{Few-Shot and Zero-Shot Learning}
\label{section2_subsec4:fewshot zeroshot}
Few-shot learning enables models to adapt to new categories using only a small number of labeled examples, while zero-shot learning (ZSL) aims to recognize unseen classes without direct supervision, leveraging knowledge learned from a separate set of base classes. Conventional ZSL approaches often map image representations to auxiliary semantic spaces constructed from human-annotated attributes or textual descriptors. While such methods offer a practical solution under constrained labeling, these methods tend to overfit to seen classes and are constrained by static, manually curated side information.

With the emergence of VLM-based approaches~\cite{radford2021learning, jia2021scaling}, zero-shot inference has improved through aligned visual-textual embeddings.
On top of these foundations, prompting-based methods~\cite{rao2022denseclip, zhang2021tip} have refined prompt-feature interaction to enhance adaptability.
However, most still rely on fixed prompt structures and static features, limiting their capacity to generalize across diverse domains.

In this work, we integrate prompt learning with attribute-aware adaptation to capture intra-class variability through dynamic representations, rather than relying on predefined prompts or fixed attribute sets. This approach enables finer discrimination of subtle visual differences and more effective adaptation in both zero-shot and few-shot settings.

\section{Methodology}
\label{section3:methodology}

\subsection{Preliminaries}
\label{section3_subsec1:preliminaries}
CLIP~\cite{radford2021learning} is built upon a dual-encoder architecture, using a ResNet~\cite{he2016deep} or ViT~\cite{dosovitskiy2020image} for image encoding and a Transformer~\cite{vaswani2017attention} for text encoding. Both encoders project inputs into a shared embedding space and are optimized using contrastive learning, which brings paired image-text features closer while pushing apart mismatched pairs.

Given an input image $x$, the image encoder $f(\cdot)$ produces a visual representation $f(x)$. To construct text embeddings, a template such as ``a photo of a \{\{class\}\}'' is filled with each class name, yielding $K$ textual prototypes $\{w_i\}_{i=1}^{K}$. Classification is then performed based on the softmax over cosine similarities:
\begin{equation}
    p(y|x) = \frac{\exp(\mathrm{sim}(f(x), w_y)/\tau)}{\sum_{i=1}^{K} \exp(\mathrm{sim}(f(x), w_i)/\tau)},
    \label{eq:cocop_contrastive_loss}
\end{equation}
where $\mathrm{sim}(\cdot, \cdot)$ denotes cosine similarity, and $\tau$ is a temperature scaling factor.

CoOp~\cite{zhou2022learning} replaces the static prompt templates with $M$ trainable context vectors $\{v_1, \dots, v_M\}$. For each class $i$, the prompt is constructed as $t_i = \{v_1, \dots, v_M, c_i\}$, where $c_i$ is the embedding of the class label. The prompt $t_i$ is then passed through CLIP’s frozen text encoder $g(\cdot)$ to obtain the class feature used for prediction.

CoCoOp~\cite{zhou2022conditional} extends this idea by adapting context tokens based on the input instance. A meta-network $h_\theta(\cdot)$ computes a conditioning vector $\pi = h_\theta(f(x))$ from the image feature, which adjusts each context token as $v_m(x) = v_m + \pi$. This results in an instance-aware prompt: \[t_i(x) = \{v_1(x), \dots, v_M(x), c_i\}.\]
The final prediction probability becomes:
\begin{equation}
    p(y|x) = \frac{\exp(\mathrm{sim}(f(x), g(t_y(x)))/\tau)}{\sum_{i=1}^{K} \exp(\mathrm{sim}(f(x), g(t_i(x)))/\tau)}.
    \label{eq:cococop_contrastive_loss}
\end{equation}
By jointly training the meta-network and context tokens, CoCoOp enables input-dependent prompt adaptation, improving generalization to unseen categories.

\begin{figure*}[t]
  \centering
  \includegraphics[width=0.95\textwidth]{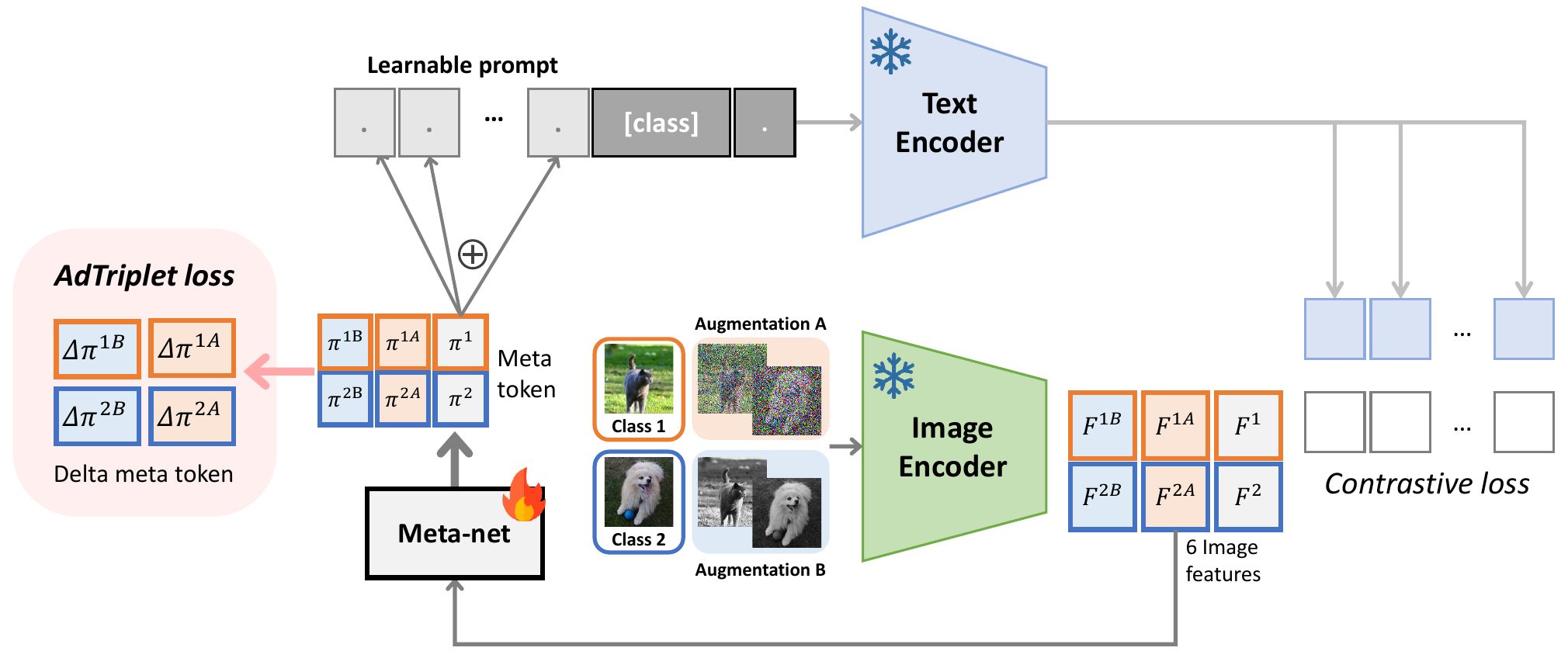}    
  \caption{\textbf{Overview of AAPL.} Two randomly augmented images are processed through a \textit{metanet} to generate meta tokens. \textit{Delta meta tokens} are then computed by subtracting class-wise means and trained with AdTriplet loss to decouple augmentation-induced attributes from class-level semantics. In parallel, learnable prompt tokens are optimized through contrastive learning between image features and class-conditioned text embeddings. The class-relevant information extracted from the \textit{meta tokens} is integrated into the prompt, allowing the model to account for augmentation-specific variation while preserving semantic consistency.}
  \label{fig:overview_figure}
\end{figure*}

\subsection{Delta Meta Token}
\label{section3_subsec2:delta meta token}
\begin{figure}[t]
    \centering
    \includegraphics[width=0.7\linewidth]{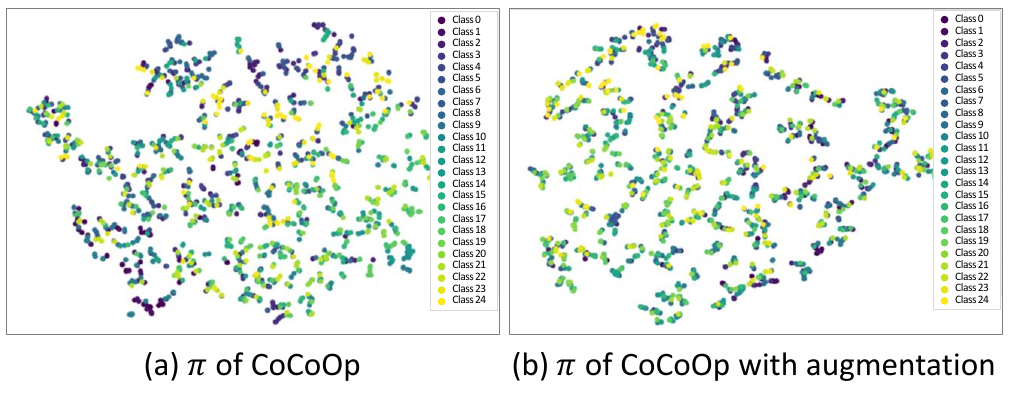}
    \caption{The comparison between \textit{meta tokens} of CoCoOp and \textit{meta tokens} of CoCoOp with random augmentation for FGVCAircraft dataset.} 
    \label{fig:aug_cocoop_figure}
\end{figure}

To investigate the effect of augmentation in prompt learning, we conducted a comparative experiment by adapting augmentation into CoCoOp~\cite{zhou2022conditional}. 
We added a conditional bias from augmented images to the learnable prompt while maintaining other settings identical to CoCoOp. 
As shown in Table~\ref{table:aug_cocoop}, incorporating augmentation reduces base-to-new generalization accuracy compared to the original CoCoOp since the \textit{metanet} fails to extract the semantic features from the augmented images, thereby capturing arbitrary noise rather than attribute-specific semantics. 
Additionally, as shown in Fig.~\ref{fig:aug_cocoop_figure}, it does not show a big difference in class clustering, indicating that the \textit{meta token} fails to capture the crucial semantic features for the classification. Consequently, this suggests that merely using augmentation in prompt learning might not enhance robustness or performance. It potentially leads to detrimental effects due to the \textit{metanet}'s inability to identify meaningful semantic features from the augmented images, focusing on instance-specific features rather than class semantics. 
To achieve optimal results, augmentation needs to be applied more carefully, ensuring that the conditional biases appropriately capture the semantic information of the class.

\begin{table}[t]
\centering
\small
\footnotesize
\renewcommand{\arraystretch}{1.1}
\setlength{\tabcolsep}{6pt}
\begin{tabular}{lccc}
\toprule
\textbf{Method} & \textbf{Base} & \textbf{New} & \textbf{HM} \\
\midrule
CoOp~\cite{zhou2022learning}           & \textbf{82.69} & 63.22 & 71.60 \\
CoCoOp~\cite{zhou2022conditional}      & 80.47 & 71.69 & 75.83 \\
CoCoOp w/ augmentation                 & 79.25 & 70.89 & 74.38 \\
\rowcolor{gray!10}
\textbf{AAPL}                 & 80.65 & \textbf{72.33} & \textbf{76.26} \\
\bottomrule
\end{tabular}
\caption{\textbf{Base-to-new generalization comparison.}  
We compare CoOp, CoCoOp, CoCoOp with augmentation, and AAPL in terms of harmonic mean (HM) accuracy.}
\label{table:aug_cocoop}
\end{table}

CoCoOp~\cite{zhou2022conditional} improves the generalization performance of CoOp~\cite{zhou2022learning} by introducing \textit{metanet}, which outputs \textit{meta token} from image samples and then adds it to the learnable prompt. It focuses on learning instance-specific information rather than class-level information. However, it's still unclear what information the \textit{meta token} contains, as the \textit{metanet} is a black box, and its shallow architecture leads to uncertain feature extraction. As shown in Fig.~\ref{fig:aug_cocoop_figure}, it fails to demonstrate clear clustering by either augmentation type or class. It shows that the \textit{meta token} does not effectively capture either the semantic information of the class or the attribute of the input image. To address this issue and make it possible to add desired information to the learnable prompt, we propose the concept of a \textit{delta meta token}, the attribute-specific bias. The overview of AAPL is shown in Fig.~\ref{fig:overview_figure}.

To make a \textit{delta meta token}, two images from each of the two different classes are required, \eg class~1 and class~2, as shown in Fig.~\ref{fig:overview_figure}. 
Two different augmentation types are randomly selected from 14 augmentations proposed in SimCLR~\cite{chen2020simple} for each pair of input images without any duplication, denoted as $Aug_A(\cdot)$ and $Aug_B(\cdot)$. 
Inspired by TextManiA~\cite{ye2023textmania}, which demonstrated the extraction of attribute information from text using Word Vector Analogy~\cite{ethayarajh2018towards, mikolov2013linguistic}, we generate \textit{delta meta token} by subtracting image features in the same class with different augmentations. \textit{Delta meta token} represents a difference vector from image features that contain augmentation information. They are generated at each iteration. The \textit{delta meta token} from an image $x$ of class 1 and $Aug_A(\cdot)$ can be written as follows:
\begin{equation}
    \Delta\pi^{1A} = h_{\theta}(f(Aug_A(x_1))) - h_{\theta}(f(x_1)) \label{eq:delta_meta_1A}.\\ 
\end{equation}

As demonstrated in TextManiA, semantic attributes derived from class-level information can effectively improve classification performance. Building on this idea, we propose a prompt learning framework that combines class semantics with augmentation-aware signals. Central to our design is the \textit{delta meta token}, which retains fine-grained visual features introduced by augmentations by incorporating decomposed auxiliary information into the prompt space. Unlike the conventional \textit{meta token}, it preserves class and attribute cues as separate components, enabling the model to capture subtle intra-class variation and improve domain generalization.

To further strengthen this mechanism, we introduce adversarial token embeddings that decouple augmentation-related appearance shifts from class-relevant semantics, reducing superficial noise in the representation. We also propose an adversarial triplet loss, AdTriplet, which modulates the prompt’s conditional bias by aligning it with consistent class semantics across views. Augmentation profiling empirically verifies that this strategy suppresses noisy attribute signals and enhances semantic coherence. Inspired by adversarial prompt learning in NLP~\cite{wu2022adversarial, nookala2023adversarial}, our approach leverages a dynamic interaction between class and attribute pathways via the \textit{metanet}, leading to more robust prompt representations that generalize across diverse visual distributions.

\begin{figure*}
    \centering
    \includegraphics[width=1.0\linewidth]{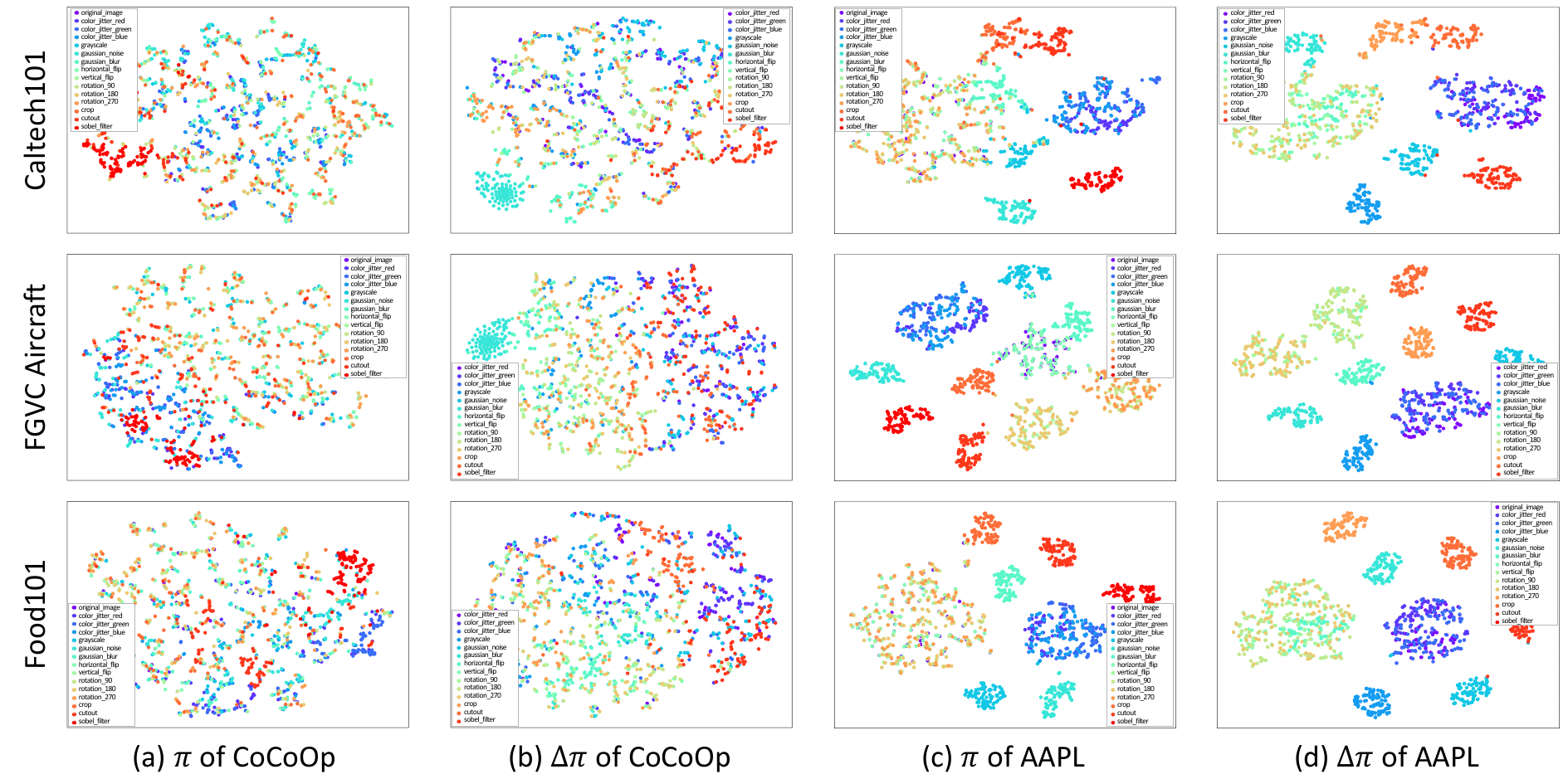}
    \caption{t-SNE visualization of the \textit{meta token} and \textit{delta meta token} from CoCoOp~\cite{zhou2022conditional} and AAPL on the Caltech101, FGVCAircraft, Food101 datasets. 
    Point colors represent the 14 different augmentations, and 100 validation samples are used for visualization. $(a)$ and $(c)$ show the \textit{meta token}, while $(b)$ and $(d)$ show the \textit{delta meta token}.}
    \label{fig:tsne_results}
\end{figure*}

In Fig.~\ref{fig:tsne_results}, we visualize the validation results of the \textit{metanet} from both CoCoOp~\cite{zhou2022conditional} and AAPL using t-SNE. The figure shows that CoCoOp struggles to distinguish between various augmentations, whereas AAPL demonstrates clearer separation. In Fig.~\ref{fig:tsne_results} (c) and (d), the \textit{meta token} fails to fully distinguish the 14 augmentation types, whereas the \textit{delta meta token} achieves near-complete separation, except for minor overlaps in cases like vertical flips and rotations. These results indicate that the \textit{delta meta token} captures augmentation-specific information better than the \textit{meta token}.

To further improve augmentation utilization, we impose an adversarial loss on the \textit{delta meta token}, restricting the \textit{metanet}’s role to classification alone. This constraint helps clarify why the feature learning mechanism in CoCoOp leads to performance gains. By refining how augmentation-driven features are decomposed and learned, we propose a more targeted approach to leveraging augmentation in prompt learning. Additionally, we conduct systematic profiling to analyze the impact of modifying decomposed feature components, providing a deeper understanding of how augmentation influences image feature selection in prompt-based learning.

Inspired by TextManiA~\cite{ye2023textmania}, which demonstrated that subtracting textual features preserves specific attributes, we extend this idea to visual features. Our findings show that the \textit{delta meta token} effectively encodes augmentation information, resulting in more precise feature separation. To the best of our knowledge, this is the first study to investigate visual feature decomposition using subtraction within prompt learning. Notably, while the \textit{meta token} still retains class-related information, the \textit{delta meta token} successfully decouples semantic content from augmentation-specific attributes.

\subsection{Adversarial Triplet Loss}
\label{section3_subsec3:adversarial triplet loss}
Using triplet loss~\cite{hoffer2015deep, sohn2016improved, weinberger2009distance, schroff2015facenet}, we can eliminate the remaining class-specific information in the \textit{delta meta token} while enhancing information related to augmentations. Training is conducted with four \textit{delta meta tokens}, \eg $\Delta\pi^{1A}$, $\Delta\pi^{1B}$, $\Delta\pi^{2A}$, and $\Delta\pi^{2B}$, in the embedding space, aiming to increase the distance between vectors of the same class while minimizing it for the same augmentation. 

\begin{figure}[t]
    \centering
    \includegraphics[width=0.65\linewidth]{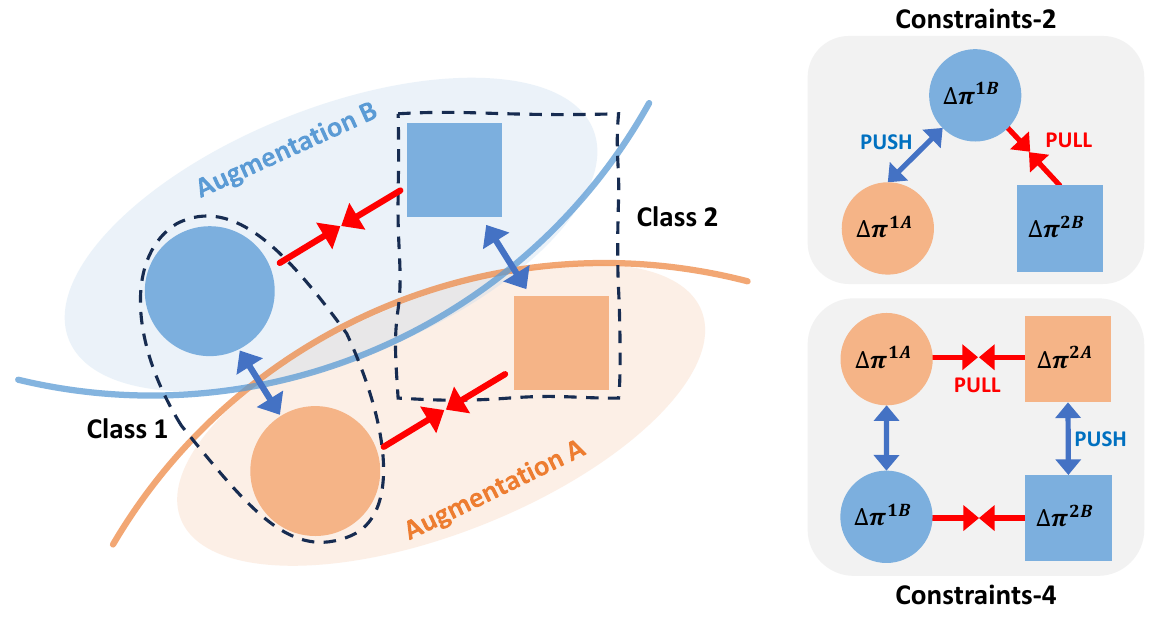}
    \caption{Comparison of the number of constraints of the AdTriplet loss. The constraints-2 setting's anchor is just one, \eg $\Delta\pi^{1B}$, and the constraints-4 setting has two anchors, \eg $\Delta\pi^{1A}$ and $\Delta\pi^{2B}$.}
    \label{fig:fig5}
\end{figure}

Fig.~\ref{fig:fig5} provides a conceptual visualization of the AdTriplet loss, showing how different constraint settings (constraints-2 and constraints-4) affect the relationships among \textit{delta meta tokens} in the embedding space. In the constraints-2 setting, the model uses a single anchor, like $\Delta\pi^{1B}$, forming one triplet and results in a relatively limited learning signal. In contrast, the constraints-4 setting employs two anchors, $\Delta\pi^{1A}$ and $\Delta\pi^{2B}$, enabling a greater variety of class-augmentation combinations. This structure allows the \textit{delta meta token} to retain augmentation-specific features, while aligning the conditional bias in prompt learning more closely with class-level semantics, thereby supporting more robust and generalizable prompt learning.

For example, if the anchor is $\Delta\pi^{1A}$, its positive pair is $\Delta\pi^{2A}$, which has a different class but the same augmentation. In contrast, $\Delta\pi^{1B}$ is a negative pair because it has the same class but a different augmentation.
The distance between the anchor and the negative pair should be greater than the distance between the anchor and the positive pair. The Euclidean distance is denoted as $\|\ {\cdot}\|_2$, and the margin of the triplet loss is denoted as $m$ in Eq.~\ref{eq:triplet_loss1}.
\begin{align}
L_{triplet}(x, x^{+}, x^{-} ;\Delta\pi^{1A}, \Delta\pi^{2A},\Delta\pi^{1B}) 
  &= \max \left(0,\ \|x - x^{+} \|_2 - \|x - x^{-}\|_2 + m \right) \notag \\
  &= \max \left(0,\ \|\Delta\pi^{1A} - \Delta\pi^{2A}\|_2 - \|\Delta\pi^{1A} - \Delta\pi^{1B}\|_2 + m \right)
  \label{eq:triplet_loss1}
\end{align}

Thus, we introduce the Adtriplet loss, which adversarially trains the model to prioritize the alignment of augmentation information over class information. This loss is optimized jointly with the classification loss, specifically the cross-entropy loss.
In our setting, the AdTriplet loss is applied in the constraints-4 configuration, as illustrated in Fig.~\ref{fig:fig5}, to maintain a balanced connection between the class information domain and augmentation attribute domain.
\begin{equation}
    L_{AdTriplet} = L_{triplet}^1(\Delta\pi^{1A}, \Delta\pi^{2A},\Delta\pi^{1B}) + L_{triplet}^2(\Delta\pi^{2B}, \Delta\pi^{1B}, \Delta\pi^{2A}) 
    \label{eq:adtriplet_loss}     
\end{equation}

The cross-entropy loss is computed following the same method as CoCoOp~\cite{zhou2022conditional}. 
To ensure fairness between the training and test phases, only one input image label is used in the cross-entropy loss calculation. The overall training loss function is as follows: 
\begin{equation}
    L_{total} = \alpha * L_{AdTriplet} + \beta * L_{CE},
\label{eq:total_loss}
\end{equation} 
where $\alpha$ and $\beta$ are hyper-parameters controlling the relative contributions of each loss term.
Detailed parameter tuning is provided in Sec.~\ref{section4:experiment}.

\section{Experiments}
\label{section4:experiment}

\subsection{Experimental Settings}
\label{section4_subsec1}
\subsubsection{Datasets}
\label{section4_subsec1_subsubsec1}
We use 11 classification datasets based on CLIP~\cite{radford2021learning}, CoOp~\cite{zhou2022learning}, and CoCoOp~\cite{zhou2022conditional} for base-to-new generalization and cross-dataset transfer: ImageNet~\cite{deng2009imagenet} and Caltech101~\cite{fei2004learning} for generic object classification, OxfordPets~\cite{parkhi2012cats}, StanfordCars~\cite{krause20133d}, Flowers102~\cite{nilsback2008automated}, Food101~\cite{bossard2014food} and FGVCAircraft~\cite{maji2013fine} for fine-grained image recognition, EuroSAT~\cite{helber2019eurosat} for satellite image classification, UCF101~\cite{soomro2012ucf101} for action classification, DTD~\cite{cimpoi2014describing} for texture classification, and SUN397~\cite{xiao2010sun} for scene recognition. 
For domain generalization experiments, we use ImageNet~\cite{deng2009imagenet} as the source dataset and 4 other ImageNet-based datasets, \ie ImageNetV2~\cite{recht2019imagenet}, ImageNetSketch~\cite{wang2019learning}, ImageNet-A~\cite{hendrycks2021natural}, and ImageNet-R~\cite{hendrycks2021many}, as the target datasets, which each contain a different kind of domain shift. 

\subsubsection{Baselines}
\label{section4_subsec1_subsubsec2}
We mainly compare AAPL with three baseline methods: the zero-shot CLIP~\cite{radford2021learning}, CoOp~\cite{zhou2022learning}, and CoCoOp~\cite{zhou2022conditional}. 
CLIP uses the hand-crafted template ``a photo of a \{class\}'' to generate the prompts for knowledge transfer.
CoOp learns a static prompt that replaces the hand-crafted prompts with learnable vectors.
CoCoOp generates dynamic prompts by adding the image-conditional prompts to the learnable prompts in CoOp.

In addition, we compare AAPL with four other prompt learning methods: ProGrad~\cite{zhu2023prompt}, KgCoOp~\cite{yao2023visual}, DiMPLe~\cite{rahman2025dimple}, and $A^3$~\cite{wang2025a3}.
ProGrad updates only the prompt components aligned with general knowledge, keeping the backbone fixed.
KgCoOp incorporates class-level textual cues for knowledge-guided context optimization.
DiMPLe disentangles invariant and spurious features across vision and language to improve out-of-distribution alignment. $A^3$ counters unlearnable examples via cross-modal adversarial feature alignment under few-shot prompt learning.
For $A^3$, base-to-new generalization was originally evaluated on only seven datasets, and domain generalization was not performed. Therefore, we use the results of the model applied to CoCoOp, as reported in their paper, and present them as $A^3$-CoCoOp in Table~\ref{table:basetonew2}, ensuring consistency with the other baselines.

\begin{table}[t]
\centering
\renewcommand{\arraystretch}{0.9}
\scriptsize
\begin{tabular}{ll}
\hline
\textbf{Parameter} & \textbf{Value / Setting} \\
\hline
    Backbone & ViT-B/16 (pre-trained CLIP) \\
    Context length & 4 tokens \\
    Context initialization & Random initialization \\
    Parameter size & 34,816 (same as CoCoOp) \\
    N shots per class & 16 shots per class \\
    Batch size (B), Epoch (E) & B = 1, E = 10 (same as CoCoOp) \\
    Triplet loss margin $m$ & 0.2 \\
    \multirow{2}{*}{$\alpha, \beta$ }
        & $\alpha = 0.2, \beta = 1$ (in Sec.~\ref{section4_subsec2}) \\
        & $\alpha = 1, \beta = 1$ (in Sec.~\ref{section4_subsec3}--\ref{exp_subsec5}) \\
\hline
\end{tabular}
\caption{Training settings for \textbf{AAPL}.}
\label{tab:training_setup}
\end{table}

\subsubsection{Training details}
\label{section4_subsec1_subsubsec3}
Our implementation is based on the CoCoOp framework~\cite{zhou2022conditional} with a pre-trained ViT-B/16 backbone from CLIP~\cite{radford2021learning}. The overall training settings, including model configurations and key hyper-parameters, are summarized in Table~\ref{tab:training_setup}. All reported results are averaged over three runs with different random seeds. For few-shot learning, we adopt the maximum-shot setting (\ie 16 shots) used in CoOp and follow the same batch size, number of epochs, and training schedule as in CoCoOp. The triplet loss margin $m$ in Eq.~\ref{eq:triplet_loss1} is set to 0.2, and the total number of parameters in AAPL is identical to that of CoCoOp. For evaluation, we use the model from the last epoch.

\subsection{Generalization from Base-to-New Classes}
\label{section4_subsec2}
We split classes evenly into base and new (\ie unseen) sets, following CoCoOp~\cite{zhou2022conditional}. All learning-based methods are trained solely on base classes. In few-shot learning, the model is evaluated with the base classes, whereas in zero-shot learning, it is evaluated with the new classes to test the model's generalizability.
In this task, we set hyper-parameters $\alpha$ and $\beta$ to 0.2 and 1. Table~\ref{table:basetonew} presents the performance results of AAPL compared to the baseline.
AAPL outperformed in 7 out of 11 datasets, with the harmonic mean of total dataset accuracy exceeding that of CoCoOp. However, performance on the DTD~\cite{cimpoi2014describing} was significantly lower. 
This is plausibly explained by the observation that geometrical augmentations, especially flips and rotations, have a limited effect on texture datasets, where they do not meaningfully change the visual patterns. This demonstrates that the effectiveness of AAPL varies across different datasets. 

\begin{table}[t]
\centering
\renewcommand{\arraystretch}{1.1}
\resizebox{\linewidth}{!}{
\begin{tabular}{llccccc @{\hskip 10pt}!{\vrule width 0.6pt} @{\hskip 10pt} llccccc}
    \toprule
    \toprule
    \textbf{Dataset} & & CLIP & CoOp & CoCoOp & \textbf{AAPL} & \textbf{$\Delta$}
    & \textbf{Dataset} & & CLIP & CoOp & CoCoOp & \textbf{AAPL} & \textbf{$\Delta$} \\
    \midrule
    \multirow{3}{*}{\shortstack{\textbf{Average on} \\ \textbf{11 datasets}}}
    & Base  & 69.34 & \textbf{82.69} & 80.47 & \cellcolor{gray!10}80.27 & \textcolor{blue}{-0.20}
    & \multirow{3}{*}{Caltech101}
    & Base  & 96.84 & \textbf{98.00} & 97.96 & \cellcolor{gray!10}97.87 & \textcolor{blue}{-0.09} \\
    & Novel & \textbf{74.22} & 63.22 & 71.69 & \cellcolor{gray!10}72.17 & \textcolor{red}{+0.48}
    &        & Novel & 94.00 & 89.81 & 93.81 & \cellcolor{gray!10}\textbf{95.10} & \textcolor{red}{+1.29} \\
    & HM    & 71.70 & 71.66 & 75.83 & \cellcolor{gray!10}\textbf{76.01} & \textcolor{red}{+0.18}
    &        & HM    & 95.40 & 93.73 & 95.84 & \cellcolor{gray!10}\textbf{96.46} & \textcolor{red}{+0.62} \\
    \midrule
    \multirow{3}{*}{OxfordPets}
    & Base  & 91.17 & 93.67 & 95.20 & \cellcolor{gray!10}\textbf{95.63} & \textcolor{red}{+0.43}
    & \multirow{3}{*}{Food101}
    & Base  & 90.10 & 88.33 & \textbf{90.70} & \cellcolor{gray!10}\textbf{90.70} & +0.00 \\
    & Novel & 97.26 & 95.29 & \textbf{97.69} & \cellcolor{gray!10}97.40 & \textcolor{blue}{-0.29}
    &        & Novel & 91.22 & 82.26 & 91.29 & \cellcolor{gray!10}\textbf{91.60} & \textcolor{red}{+0.31} \\
    & HM    & 94.12 & 94.47 & 96.43 & \cellcolor{gray!10}\textbf{96.51} & \textcolor{red}{+0.08}
    &        & HM    & 90.66 & 85.19 & 90.99 & \cellcolor{gray!10}\textbf{91.15} & \textcolor{red}{+0.16} \\
    \midrule
    \multirow{3}{*}{Flowers102}
    & Base  & 72.08 & \textbf{97.60} & 94.87 & \cellcolor{gray!10}95.10 & \textcolor{red}{+0.23}
    & \multirow{3}{*}{Stanford Cars}
    & Base  & 63.37 & \textbf{78.12} & 70.49 & \cellcolor{gray!10}70.33 & \textcolor{blue}{-0.16} \\
    & Novel & \textbf{77.80} & 59.67 & 71.75 & \cellcolor{gray!10}70.63 & \textcolor{blue}{-1.12}
    &        & Novel & \textbf{74.89} & 60.40 & 73.59 & \cellcolor{gray!10}73.50 & \textcolor{blue}{-0.09} \\
    & HM    & 74.83 & 74.06 & \textbf{81.71} & \cellcolor{gray!10}81.06 & \textcolor{blue}{-0.65}
    &        & HM    & 68.65 & 68.13 & \textbf{72.01} & \cellcolor{gray!10}71.88 & \textcolor{blue}{-0.13} \\
    \midrule
    \multirow{3}{*}{ImageNet}
    & Base  & 72.43 & 76.47 & 75.98 & \cellcolor{gray!10}\textbf{76.53} & \textcolor{red}{+0.55}
    & \multirow{3}{*}{SUN397}
    & Base  & 69.36 & 80.60 & \textbf{79.74} & \cellcolor{gray!10}79.65 & \textcolor{blue}{-0.09} \\
    & Novel & 68.14 & 67.88 & 70.43 & \cellcolor{gray!10}\textbf{70.57} & \textcolor{red}{+0.14}
    &        & Novel & 75.35 & 65.89 & 76.86 & \cellcolor{gray!10}\textbf{76.90} & \textcolor{red}{+0.04} \\
    & HM    & 70.22 & 71.92 & 73.10 & \cellcolor{gray!10}\textbf{73.43} & \textcolor{red}{+0.33}
    &        & HM    & 72.23 & 72.51 & \textbf{78.27} & \cellcolor{gray!10}78.25 & \textcolor{blue}{-0.02} \\
    \midrule
    \multirow{3}{*}{UCF101}
    & Base  & 70.53 & \textbf{84.69} & 82.33 & \cellcolor{gray!10}82.20 & \textcolor{blue}{-0.13}
    & \multirow{3}{*}{EuroSAT}
    & Base  & 56.48 & \textbf{92.19} & 87.49 & \cellcolor{gray!10}87.00 & \textcolor{blue}{-0.49} \\
    & Novel & \textbf{77.50} & 56.05 & 73.45 & \cellcolor{gray!10}74.27 & \textcolor{red}{+0.82}
    &        & Novel & 64.05 & 54.74 & 60.04 & \cellcolor{gray!10}\textbf{66.30} & \textcolor{red}{+6.26} \\
    & HM    & 73.85 & 67.46 & 77.64 & \cellcolor{gray!10}\textbf{78.03} & \textcolor{red}{+0.39}
    &        & HM    & 60.03 & 68.69 & 71.21 & \cellcolor{gray!10}\textbf{75.25} & \textcolor{red}{+4.04} \\
    \midrule
    \multirow{3}{*}{FGVCAircraft}
    & Base  & 27.19 & \textbf{40.44} & 33.41 & \cellcolor{gray!10}34.07 & \textcolor{red}{+0.66}
    & \multirow{3}{*}{DTD}
    & Base  & 53.24 & \textbf{79.44} & 77.01 & \cellcolor{gray!10}73.90 & \textcolor{blue}{-3.11} \\
    & Novel & \textbf{36.29} & 22.30 & 23.71 & \cellcolor{gray!10}24.17 & \textcolor{red}{+0.46}
    &        & Novel & \textbf{59.90} & 41.18 & 56.00 & \cellcolor{gray!10}53.43 & \textcolor{blue}{-2.57} \\
    & HM    & \textbf{31.09} & 28.75 & 27.74 & \cellcolor{gray!10}28.28 & \textcolor{red}{+0.54}
    &        & HM    & 56.37 & 54.24 & \textbf{64.85} & \cellcolor{gray!10}62.02 & \textcolor{blue}{-2.83} \\
    \bottomrule
    \bottomrule
    \end{tabular}}
\caption{\textbf{Base-to-new generalization} with 16-shot training on base classes and evaluation on novel classes. HM denotes the harmonic mean. $\Delta$ represents the performance gap between AAPL and CoCoOp. The highest score in each column is in \textbf{bold}.
}
\label{table:basetonew}
\end{table}

\begin{table}[ht]
\centering
\scriptsize
\setlength{\tabcolsep}{3.5pt}
\renewcommand{\arraystretch}{1.05}
\begin{tabular}{lc*{12}{c}}
    \toprule
    Method & \textbf{Avg.} &
    \rotatebox{90}{ImageNet} &
    \rotatebox{90}{Caltech101} &
    \rotatebox{90}{OxfordPets} &
    \rotatebox{90}{StanfordCars} &
    \rotatebox{90}{Flowers102} &
    \rotatebox{90}{Food101} &
    \rotatebox{90}{FGVCAircraft} &
    \rotatebox{90}{SUN397} &
    \rotatebox{90}{DTD} &
    \rotatebox{90}{EuroSAT} &
    \rotatebox{90}{UCF101} \\
    \midrule
    \textbf{CoCoOp}   & 75.83 & 73.10 & 95.84 & 96.43 & 72.01 & 81.71 & 90.99 & 27.74 & 78.27 & \textbf{64.85} & 71.21 & 77.64 \\
    \textbf{Prograd}  & 76.16 & 71.46 & 95.91 & 96.33 & 72.88 & 82.03 & 89.98 & 32.82 & 77.55 & 62.45 & 72.67 & 79.35 \\
    \textbf{KgCoOp}   & \textbf{77.00} & 72.78 & 96.03 & 96.18 & \textbf{73.36} & \textbf{83.65} & 91.09 & \textbf{34.83} & \textbf{78.36} & 64.35 & 73.48 & \textbf{79.65} \\
    \textbf{DiMPLe}  & 74.70 & 66.96 & 95.44 & 94.55 & 70.89 & 79.39 & 90.51 & 33.4 & 74.85 & 65.84 & 71.11 & 76.69 \\
    \textbf{${A^3}$-CoCoOp}  & \textendash & 73.09 & 95.73 & 93.63 & \textendash & 79.74 & 90.98 & \textendash & 76.14 & \textendash & \textendash & 75.88 \\
    \rowcolor{gray!10}
    \textbf{AAPL}  & 76.01 & \textbf{73.43} & \textbf{96.46} & \textbf{96.51} & 71.88 & 81.06 & \textbf{91.15} & 28.28 & 78.25 & 62.02 & \textbf{75.25} & 78.03 \\
    \bottomrule
\end{tabular}
\caption{\textbf{Base-to-new generalization comparison} HM accuracy between CoCoOp~\cite{zhou2022conditional}, Prograd~\cite{zhu2023prompt}, KgCoOp~\cite{yao2023visual}, DiMPLe~\cite{rahman2025dimple}, $A^3$-CoCoOp ($A^3$ applied CoCoOp)~\cite{wang2025a3} and AAPL, with 16-shot training. `\textendash' indicates that the metric is not reported in the original paper.}
\label{table:basetonew2}
\end{table}

\begin{figure}[!h]
    \centering
    \includegraphics[width=0.85\linewidth]{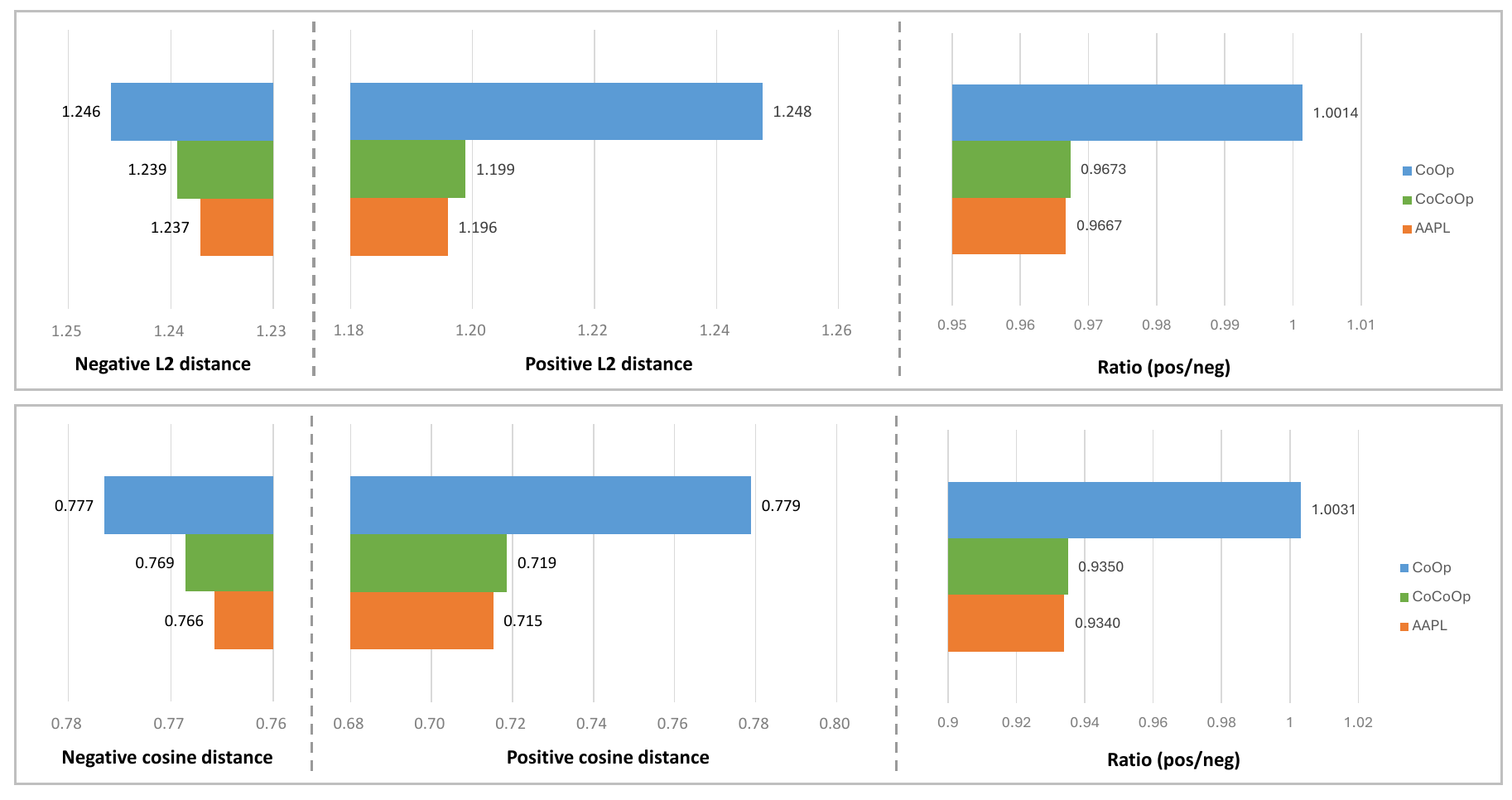}
    \caption{\textbf{Comparison of prompt distance metrics}, \ie \textit{L2} and cosine distance, among CoOp, CoCoOp, and AAPL on FGVCAircraft dataset. The positive distance refers to the similarity between the input and its ground-truth prompt, while the negative distance represents the average distance to all other class prompts. Lower Positive and higher Negative indicate better alignment and separation.}
    \label{fig:prompt_img}
\end{figure}

To further validate generality, we additionally compare with ProGrad~\cite{zhu2023prompt}, KgCoOp~\cite{yao2023visual}, DimPLe~\cite{rahman2025dimple}, and $A^3$~\cite{wang2025a3}, which adopt different prompt learning paradigms. 
As shown in Table~\ref{table:basetonew2}, AAPL achieves competitive performance across most datasets, often ranking first or second, while maintaining robustness against diverse and recent prompt tuning strategies. 
This indicates that the advantages of AAPL are not restricted to a specific baseline design but extend to varied prompt learning frameworks.

In addition, we measured the computational overhead relative to CoCoOp.
AAPL requires about 1.25× longer training time due to the added steps of augmentation profiling and adversarial optimization, yet its inference speed remains nearly identical (1.01×).
This shows that the observed performance gains are achieved with only a modest increase in training cost and negligible impact on practical usability.

To probe model behavior, we measure positive and negative distances on FGVCAircraft, as illustrated in Fig.~\ref{fig:prompt_img}. The positive distance is the distance between an image feature and its ground-truth prompt, and the negative distance is the mean distance to all other class prompts. Cosine distance is reported as $\textit{1 - cosine similarity}$ for consistency with $L2$ distance; lower positive and higher negative are better. 
Among the baselines, AAPL attains the smallest positive, which means a strong image-prompt alignment, but also a smaller negative distance than the others. We hypothesize that attribute-guided augmentations tighten within-class consistency while shrinking inter-class margins, yielding a compact yet less separable embedding space.

\begin{figure}[!t]
    \centering
    \includegraphics[width=0.8\linewidth]{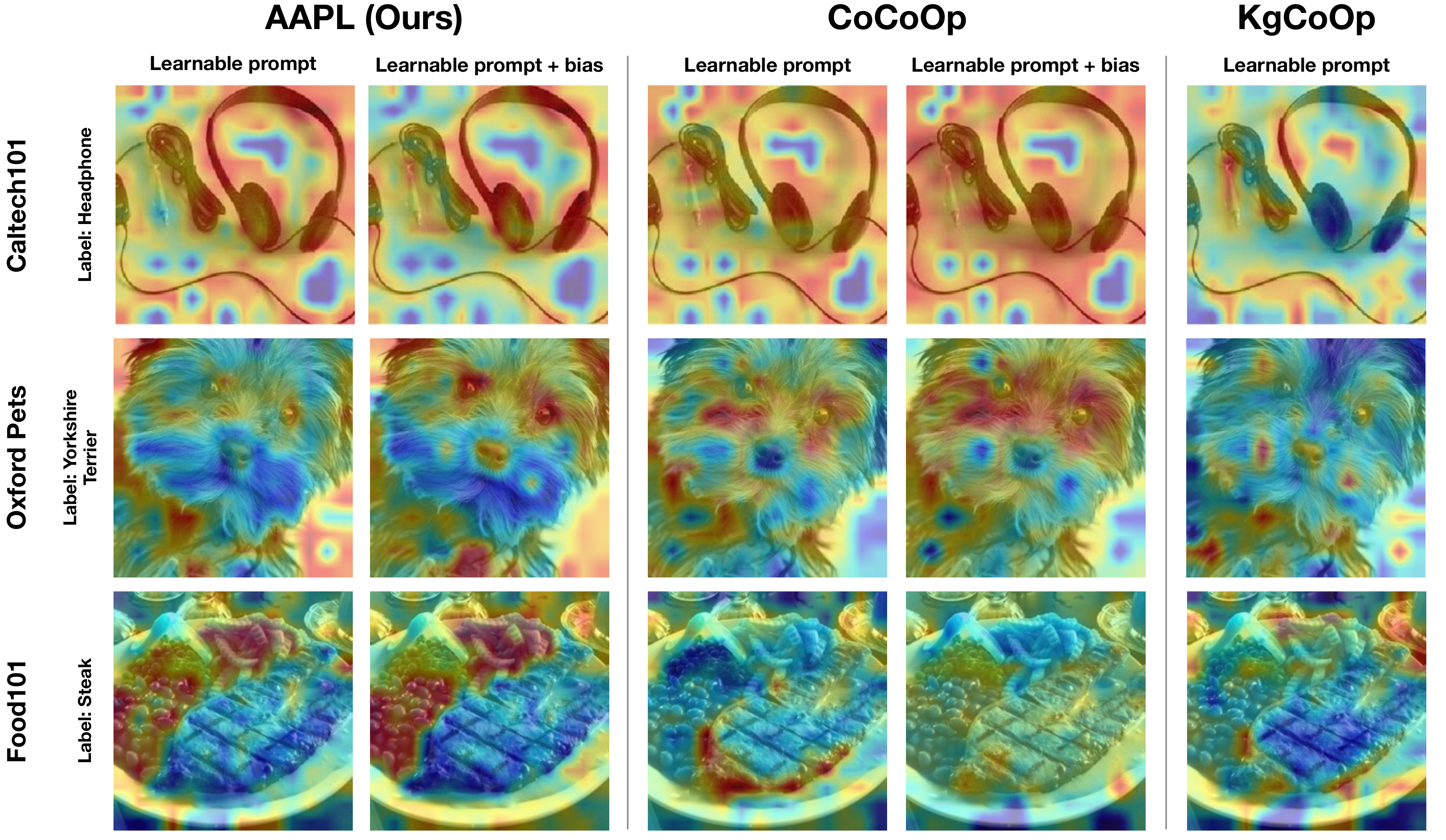}
    \caption{\textbf{Qualitative comparison of visual saliency maps} computed from cosine similarity between normalized patch and prompt features, scaled to [0,1] for visualization~\cite{pmlr-v235-zhao24p}. Comparisons are shown for \textbf{CoCoOp~\cite{zhou2022conditional}}, \textbf{AAPL}, and \textbf{KgCoOp~\cite{yao2023visual}} on \textit{Caltech101}, \textit{OxfordPets}, and \textit{Food101}. 
    Columns present: AAPL with \emph{learnable prompt} and \emph{learnable prompt + delta meta token learned-bias}, CoCoOp with \emph{learnable prompt} and \emph{learnable prompt + meta token bias}, and KgCoOp with \emph{learnable prompt}.}
    \label{fig:attn_maps}
\end{figure}

Fig.~\ref{fig:attn_maps} compares visual saliency maps computed using cosine similarity with min-max normalization between image and learnable prompt features across CoCoOp, AAPL, and KgCoOp on Caltech101, Oxford Pets, and Food101. Unlike softmax-based attention maps, this approach preserves the absolute activation strength of each image region by directly measuring cosine similarity between normalized patch and prompt embeddings, followed by min-max scaling for visualization. CoCoOp’s \textit{meta token} bias provides a global prior, yielding heatmaps with strong overall intensity (\ie hotter colors) but lower relative contrast on discriminative cues. In contrast, AAPL’s \textit{delta meta token} bias, learned as the feature difference $(\Delta)$ between original and augmented pairs, suppresses background/style noise, highlighting object-centric semantics by aligning prompt and image patch features.
Consequently, heatmaps focus on discriminative parts (\eg col.2: row1, headband/cable; row2, eyes/nose; row3, beans/fries), increasing contrast and relevance. KgCoOp generally shows weaker visual grounding, often failing to strongly activate the object even in object-centric datasets, suggesting greater reliance on textual than visual evidence. AAPL can also overfocus on local patterns, reducing gains for datasets that require global texture or layout understanding, such as DTD and EuroSAT.

\subsection{Cross-Dataset Transfer} 

\begin{table*}[t]
\centering
\small
\renewcommand{\arraystretch}{0.9}
\setlength{\tabcolsep}{5pt}
\resizebox{\linewidth}{!}{
\begin{tabular}{l c c c c c c c c c c c c}
\toprule
& \textbf{Source} & \multicolumn{10}{c}{\textbf{Target}} & \textbf{Average} \\
\cmidrule(lr){2-2} \cmidrule(lr){3-12} \cmidrule(lr){13-13}
  & \rotatebox{90}{ImageNet}
  & \rotatebox{90}{Caltech101} 
  & \rotatebox{90}{OxfordPets} 
  & \rotatebox{90}{StanfordCars} 
  & \rotatebox{90}{Flowers102} 
  & \rotatebox{90}{Food101} 
  & \rotatebox{90}{FGVCAircraft} 
  & \rotatebox{90}{SUN397} 
  & \rotatebox{90}{DTD} 
  & \rotatebox{90}{EuroSAT} 
  & \rotatebox{90}{UCF101} 
  & \rotatebox{90}{\textit{Avg.}} \\ 
\midrule
\textbf{CoOp}~\cite{zhou2022learning} 
& \textbf{71.51} & 93.70 & 89.14 & 64.51 & 68.71 & 85.30 & 18.47 & 64.15 & 41.92 & 46.39 & 66.55 & 67.26 \\
\textbf{CoCoOp}~\cite{zhou2022conditional} 
& 71.02 & \textbf{94.43} & 90.14 & \textbf{65.32} & \textbf{71.88} & \textbf{86.02} & 22.94 & \textbf{67.36} & \textbf{45.73} & \textbf{45.37} & 68.21 & \textbf{68.72} \\
\rowcolor{gray!10}
\textbf{AAPL} 
& 71.37 & 94.17 & \textbf{90.73} & 65.10 & 71.67 & 86.00 & \textbf{23.03} & 66.80 & 44.80 & 41.83 & \textbf{69.30} & 68.36 \\
\bottomrule
\end{tabular}
}
\caption{\textbf{Cross-dataset transfer experiment.} All models are trained on the full class set of \textbf{ImageNet (source)} and evaluated on 10 target datasets. The final column shows the average target accuracy.}
\label{table:cross_dataset}
\end{table*}

\label{section4_subsec3}
To assess the robustness and adaptability of AAPL, we evaluate its cross-dataset generalization by training on all 1,000 ImageNet classes and testing on 10 other datasets shown in Table~\ref{table:cross_dataset}. We assume that the model can capture semantic information about image features by learning precise attributes. Except for Sec.~\ref{section4_subsec1_subsubsec2}, we set both $\alpha$ and $\beta$ to 1 in all experiments. This setting encourages adversarial learning of visual attributes against the cross-entropy loss, improving adaptation to unseen domains.
AAPL achieves better generalization in 3 datasets: OxfordPets~\cite{parkhi2012cats}, FGVCAircraft~\cite{maji2013fine}, and UCF101~\cite{soomro2012ucf101}, compared to CoCoOp~\cite{zhou2022conditional}. By contrast, performance on DTD~\cite{cimpoi2014describing} and EuroSAT~\cite{helber2019eurosat} is notably worse. This suggests that AAPL’s augmentation-based prompt learning is less effective for datasets dominated by global properties, textures and long-range satellite scenes, rather than object-centric cues. Consequently, extracting specific attributes from these datasets is more challenging.

\subsection{Domain Generalization}
\label{section4_subsec4}
For domain generalization, we trained our model on the full ImageNet dataset, as in Sec.~\ref{section4_subsec3}, and evaluated it on four datasets representing domain shifts from ImageNet (\eg ImageNetV2~\cite{recht2019imagenet}, ImageNetSketch~\cite{wang2019learning}, ImageNet-A~\cite{hendrycks2021natural}, and ImageNet-R~\cite{hendrycks2021many}). 
We compare AAPL against CLIP, CoOp, CoCoOp, and additional prompt learning methods, ProGrad~\cite{zhu2023prompt}, KgCoOp~\cite{yao2023visual}, and DiMPLe~\cite{rahman2025dimple}. 
As shown in Table~\ref{table:domain_generalization}, AAPL achieves the highest average accuracy and outperforms all methods on ImageNet-R, while remaining competitive on the other domain-shifted datasets. 
In particular, AAPL exceeds CoCoOp in accuracy on most datasets, demonstrating that the attribute-specific bias in AAPL effectively handles domain shift and remains robust.

\begin{table*}[!t]
\centering
\scriptsize
\renewcommand{\arraystretch}{0.85}
\setlength{\tabcolsep}{5pt}
\begin{tabular}{l c c c c c c}
\toprule
& \textbf{Source} & \multicolumn{5}{c}{\textbf{Target (domain shifted)}} \\
\cmidrule(lr){3-7}
& ImageNet & IN-V2 & IN-S & IN-A & IN-R & \textit{Avg.} \\
\midrule
CLIP & 66.73 & 60.83 & 46.15 & 47.77 & 73.96 & 57.18 \\
CoOp & 71.51 & 64.20 & 47.99 & 49.71 & 75.21 & 59.28 \\
CoCoOp & 71.02 & 64.07 & 48.75 & 50.63 & 76.18 & 59.91 \\
Prograd & \textbf{72.24} & \textbf{64.73} & 47.61 & 49.39 & 74.58 & 59.08 \\
KgCoOp & 71.20 & 64.10 & \textbf{48.97} & \textbf{50.69} & 76.70 & 60.11 \\
DiMPLe & 69.73 & 61.2 & 45.67 & 44.07 & 73.87 & 58.91 \\
\rowcolor{gray!10}
\textbf{AAPL} & 71.37 & 64.20 & 48.80 & 50.60 & \textbf{76.87} & \textbf{60.12} \\
\bottomrule
\end{tabular}
\caption{\textbf{Domain Generalization experiment.} Models are trained on ImageNet (16-shot) and evaluated on four domain-shifted versions: ImageNet-V2, -S, -A, -R.}
\label{table:domain_generalization}
\end{table*}

\subsection{Augmentation Profiling}
\label{exp_subsec5}
\subsubsection{Why should the \textit{delta meta token} learn about attributes rather than class information?}
\label{exp_subsubsec4}

In zero-shot classification, relying solely on class labels limits generalization to unseen categories, as class-specific information is discrete and closely related to the training set. In contrast, attributes such as texture, color, and shape are shared across categories and provide richer semantic cues transferable to new tasks. Thus, the \textit{delta meta token} should focus on capturing attribute-level information rather than memorizing class identities.

To evaluate the effect of attribute-based learning, we measured silhouette scores~\cite{shahapure2020cluster} under different augmentation conditions.
This metric balances intra-cluster cohesion and inter-cluster separation, with higher values indicating features that are both compact and well-seperated.
Formally, for a data point $i$: $ S(i) = \frac{b(i) - a(i)}{\max\{a(i), b(i)\}} $, where $a(i)$ is the average distance to points in the same cluster, and $b(i)$ is the smallest average distance to the nearest neighboring cluster.

\begin{figure}[t]
    \centering
    \includegraphics[width=0.6\linewidth]{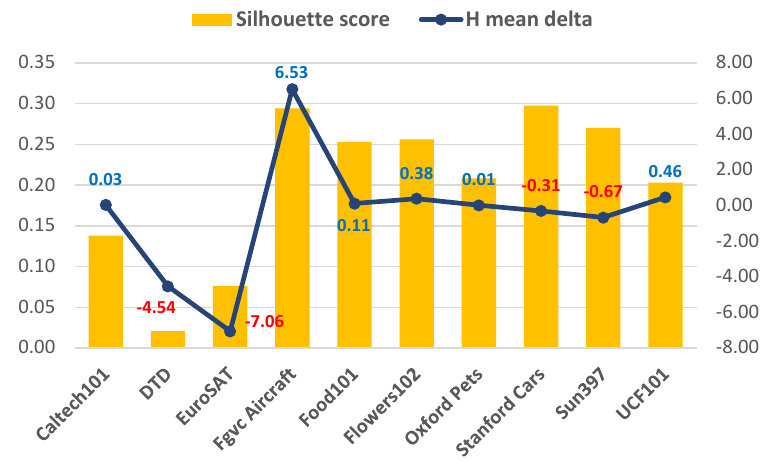}
    \caption{\textbf{The correlation between silhouette score and generalization performance.}
    Silhouette score and the difference in harmonic mean accuracy for zero-shot classification between CoCoOp and AAPL.}
    \label{fig:fig_sil_score}
\end{figure}

Using the AdTriplet loss produced more compact and distinguishable attribute representations, a trend consistent across most datasets as shown in Fig.~\ref{fig:fig_sil_score}. However, datasets such as DTD~\cite{cimpoi2014describing} and EuroSAT~\cite{helber2019eurosat} showed significant drops in both silhouette score and accuracy, revealing limitations in modeling coherent attribute-based information for texture- or layout-centric data.

\begin{table*}[!b]
\centering
\scriptsize
\renewcommand{\arraystretch}{1}
\setlength{\tabcolsep}{5pt}
\resizebox{0.9\linewidth}{!}{
\begin{tabular}{l  c c c c c c c c c c c  c}
\toprule
& \rotatebox{90}{ImageNet} 
& \rotatebox{90}{Caltech101} 
& \rotatebox{90}{OxfordPets} 
& \rotatebox{90}{StanfordCars} 
& \rotatebox{90}{Flowers102} 
& \rotatebox{90}{Food101} 
& \rotatebox{90}{FGVCAircraft} 
& \rotatebox{90}{SUN397} 
& \rotatebox{90}{DTD} 
& \rotatebox{90}{EuroSAT} 
& \rotatebox{90}{UCF101} 
& \rotatebox{90}{\textit{Average}} \\
\midrule
\textbf{Triplet} & \textbf{73.44} & 95.81 & 96.18 & \textbf{72.22} & 80.65 & 90.70 & 27.97 & \textbf{78.34} & \textbf{61.73} & 64.15 & \textbf{78.78} & 74.54 \\
\rowcolor{gray!10}
\textbf{AdTriplet} & 73.09 & \textbf{96.87} & \textbf{96.44} & 71.70 & \textbf{82.09} & \textbf{91.10} & \textbf{34.27} & 77.60 & 60.31 & \textbf{65.16} & 78.10 & \textbf{75.16} \\
\bottomrule
\end{tabular}
}
\caption{\textbf{Comparison of AAPL with Triplet vs. AdTriplet loss.} Accuracy is measured as harmonic mean in base-to-new generalization across 11 datasets.}
\label{table:triplet_vs_adtriplet}
\end{table*}

\begin{figure}[t]
    \centering
    \includegraphics[width=0.8\linewidth]{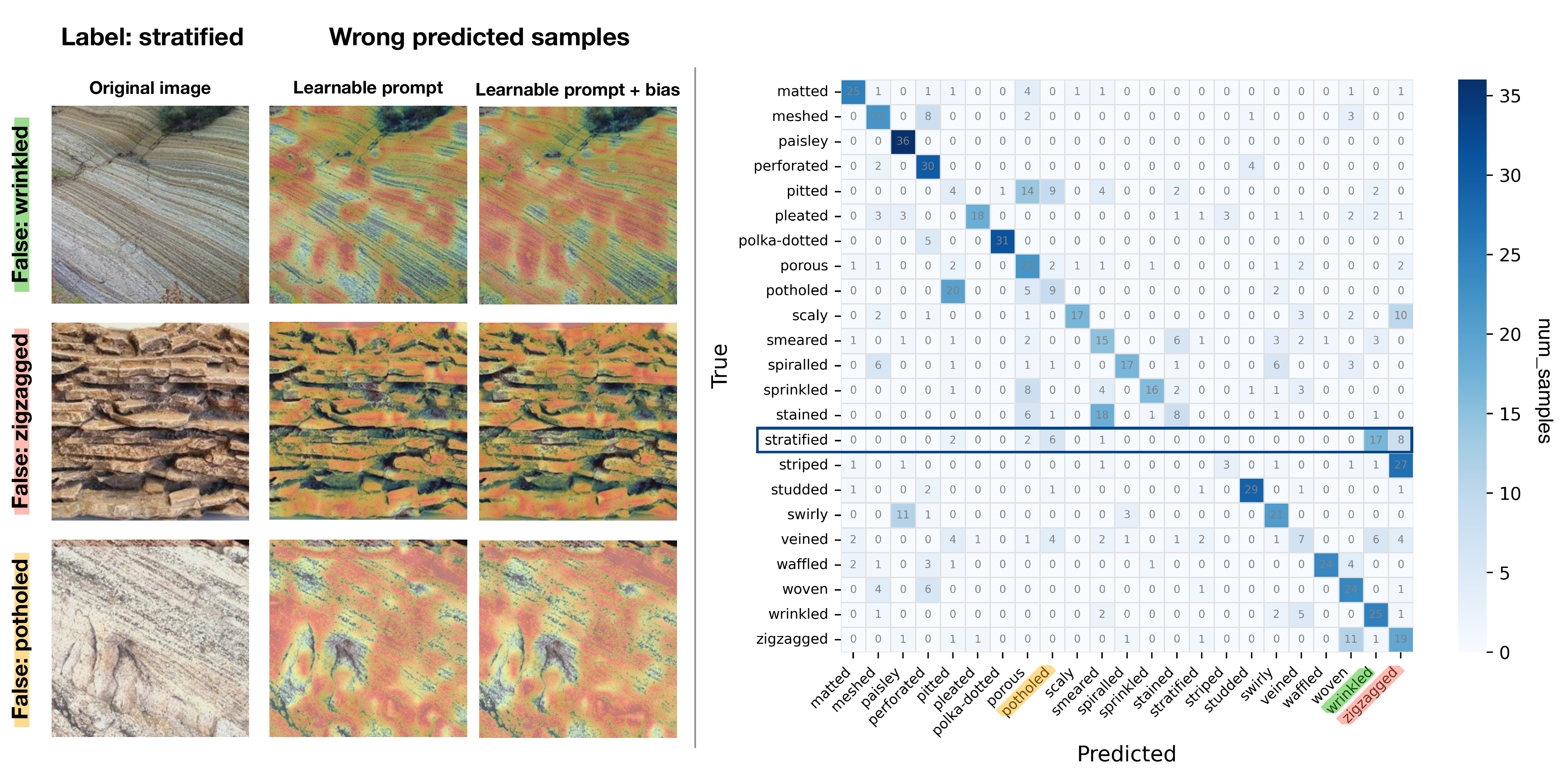}
    \caption{\textbf{DTD~\cite{cimpoi2014describing} validation with AAPL.} Right: confusion matrix. Left: for the lowest-accuracy class, \textit{stratified}, we display its three most frequent confusions, \textit{wrinkled}, \textit{zigzagged}, and \textit{potholed}, together with the input image and visual saliency maps, as in Fig.~\ref{fig:attn_maps}, using the \textit{learnable prompt} and the \textit{learnable prompt + delta meta token learned bias}. The saliency maps remain nearly unchanged after adding the bias, indicating that the proposed \textit{delta meta token} fails to redirect focus for texture-centric categories.}
    \label{fig:confusion_analysis}
\end{figure} 

\subsubsection{Which dataset is vulnerable for \texttt{\textbf{AAPL}}?}
\label{exp_subsubsec5}
We analyzed silhouette scores~\cite{shahapure2020cluster} for different augmentation types to assess how effectively the \textit{delta meta token} captures augmentation-sensitive attributes.
The AdTriplet loss encourages discrimination of fine-grained, augmentation-induced attributes while preserving class identity, whereas the traditional triplet loss clusters samples mainly by class, ignoring augmentation differences.
As presented in Table~\ref{table:triplet_vs_adtriplet}, the AdTriplet loss yielded performance improvements in most datasets, indicating that learning attribute-level distinctions supports stronger generalization in zero-shot tasks. 
However, FGVCAircraft~\cite{maji2013fine} performed about 7\% better with the triplet loss, indicating a stronger reliance on class-level structure. This suggests AAPL is more effective for datasets with diverse visual attributes, whereas conventional losses may suit class-dominant datasets.

\begin{figure}[t]
    \centering
    \includegraphics[width=0.75\linewidth]{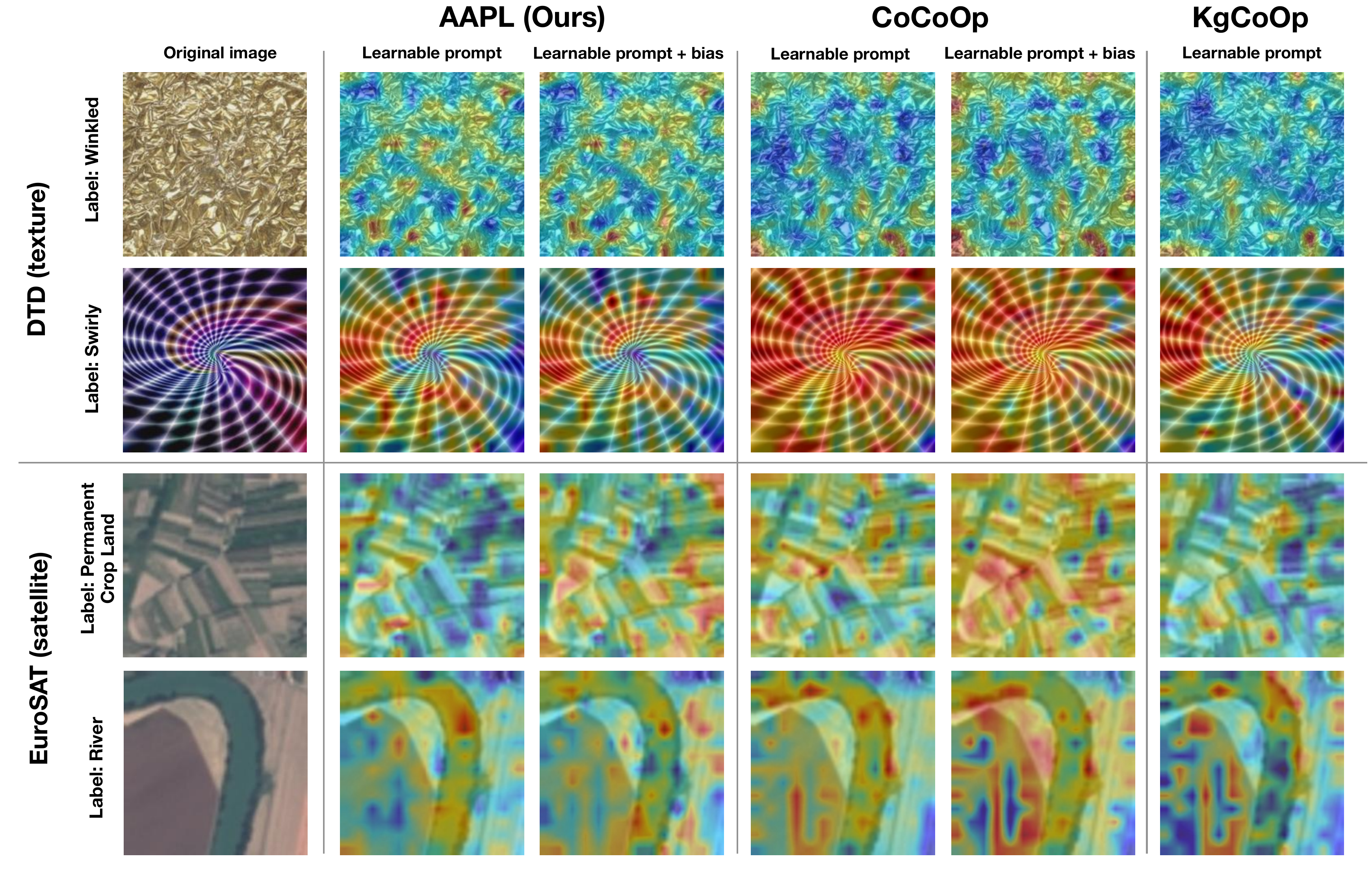}
    \caption{\textbf{Qualitative comparison of visual saliency maps on DTD and EuroSAT.}For each image, the original input is shown alongside saliency maps for \textbf{CoCoOp}, \textbf{KgCoOp}, and \textbf{AAPL}.
    Across both datasets, AAPL’s \textit{delta meta token} produces only minor changes and fails to restore broad, layout-level coverage (\ie global features), whereas CoCoOp retains comparatively wider and more global coverage.}
    \label{fig:euro_dtd_attn_maps}
\end{figure} 

As shown in Fig.~\ref{fig:confusion_analysis}, for DTD~\cite{cimpoi2014describing}, where labels are defined primarily by global texture rather than object-centric cues, AAPL offers limited benefit from the \textit{delta meta token}. In the confusion matrix, the class \textit{stratified} is often predicted as the texture-similar classes \textit{wrinkled}, \textit{zigzagged}, or \textit{potholed}. Fig.~\ref{fig:confusion_analysis} visualizes three representative failure cases in which the incorrect class attains the highest class logit, and the corresponding visual saliency lacks clear activation on discriminative texture regions of the true class, instead responding more strongly to patterns characteristic of the wrong label. Moreover, adding the \textit{delta meta token} bias to the learnable prompt yields only marginal changes in these saliency maps, suggesting that the method does not redirect focus toward discriminative global patterns on DTD. In line with this, the saliency overlays for DTD and EuroSAT in Fig.~\ref{fig:euro_dtd_attn_maps} reveal insufficient broad, scene- or layout-level coverage, whereas CoCoOp retains comparatively wider and more global coverage. By contrast, on object-centric datasets in Fig.~\ref{fig:attn_maps}, the \textit{delta meta token} more noticeably modulates saliency when added to the learnable prompt.

\subsubsection{Which augmentation is effective to prompt learning?}
\label{exp_subsubsec6}
\begin{figure}[t]
    \centering
    \includegraphics[width=1.0\linewidth]{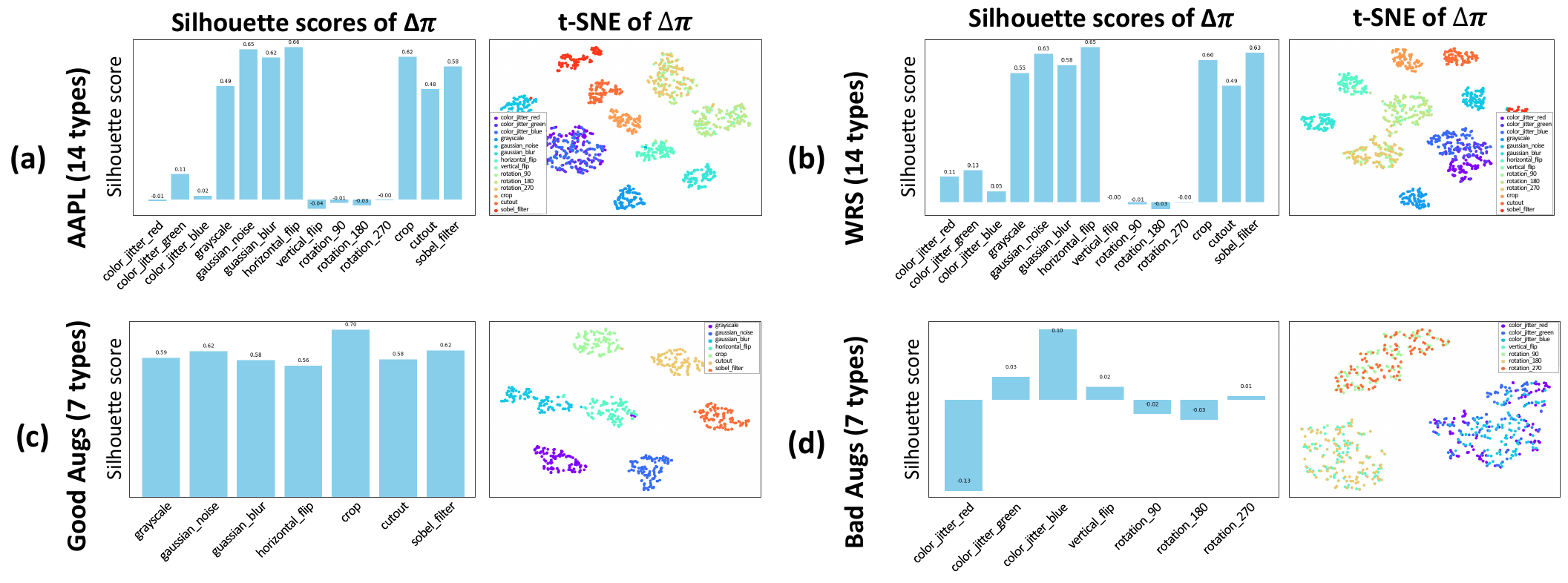}
    \caption{\textbf{The comparison of silhouette score and t-SNE} of the base-to-new generalization for each of the specific augmentation types on FGVCAircraft. All results are from the last epoch.}
    \label{fig:good_bad_aug}
\end{figure}

Extending the previous analysis, we examine which augmentations are most effective for prompt learning. We visualized the \textit{delta meta token} embeddings using t-SNE across 14 different augmentation types and computed their silhouette scores (Fig.~\ref{fig:good_bad_aug} (a)). The visualizations show that some augmentations, such as horizontal flips and rotations, as well as color jitters, result in overlapping patterns that are challenging to distinguish in the embedding space. This lack of separability was observed consistently across all datasets, suggesting that such ambiguous augmentations limit the model’s ability to form distinct attribute representations.

We trained the model using only “good augmentations” that formed clearly separable clusters (Fig.~\ref{fig:good_bad_aug} (b)), which improved base-to-new generalization, clustering quality, and silhouette scores (Table~\ref{table:good_bad_aug}). In contrast, using only “bad augmentations” (Fig.~\ref{fig:good_bad_aug} (c)) yielded no material improvement in either separation or accuracy. Bad image augmentations, \eg strong color jitters or texture distortions~\cite{willbo2024impacts}, can hinder the metanet’s ability to model augmentation-sensitive features, thereby reducing the overall generalization capability of prompt-based learning methods.

\begin{table}[t]
\centering
\scriptsize
\renewcommand{\arraystretch}{1}
\setlength{\tabcolsep}{4pt}
\begin{tabular}{l ccccccccccc c}
\toprule
\textbf{Method} &
\rotatebox{90}{ImageNet} &
\rotatebox{90}{Caltech101} &
\rotatebox{90}{OxfordPets} &
\rotatebox{90}{StanfordCars} &
\rotatebox{90}{Flowers102} &
\rotatebox{90}{Food101} &
\rotatebox{90}{FGVCAircraft} &
\rotatebox{90}{SUN397} &
\rotatebox{90}{DTD} &
\rotatebox{90}{EuroSAT} &
\rotatebox{90}{UCF101} &
\rotatebox{90}{Average} \\
\midrule
\textbf{AAPL}        & 73.09 & 95.87 & 96.44 & 71.70 & \textbf{82.09} & \textbf{91.10} & \textbf{34.27} & 77.60 & 60.31 & 64.15 & 78.10 & 74.97 \\
\textbf{Good Augs}   & 72.91 & \textbf{96.43} & 96.49 & \textbf{71.85} & 80.80 & 90.45 & 34.02 & 77.97 & 61.24 & 66.68 & 77.09 & 75.08 \\
\textbf{Bad Augs}    & 73.05 & 96.00 & 95.96 & 71.67 & 81.74 & 90.90 & 18.14 & 78.03 & \textbf{61.43} & \textbf{74.70} & 78.11 & 74.52 \\
\rowcolor{gray!10}
\textbf{WRS}         & \textbf{73.25} & 96.40 & \textbf{96.49} & 71.82 & 82.01 & 90.96 & 26.53 & \textbf{78.14} & 61.39 & 74.25 & \textbf{78.33} & \textbf{75.42} \\
\bottomrule
\end{tabular}
\caption{\textbf{Effect of augmentation type and sampling strategy in AAPL.} 
Harmonic mean accuracy for each augmentation strategy across datasets. \text
{WRS} applies weighted random sampling based on silhouette scores.}
\label{table:good_bad_aug}
\end{table}

\begin{table}[t]
\centering
\footnotesize
\renewcommand{\arraystretch}{1}
\setlength{\tabcolsep}{6pt}
\begin{tabular}{lcc>{\columncolor{gray!10}}c}
\toprule
\textbf{Dataset} & \textbf{AAPL} & \textbf{WRS} & $\boldsymbol{\Delta}$ \\
\midrule
StanfordCars & 71.70 & 71.82 & \textbf{+0.12} \\
SUN397       & 77.60 & 78.14 & \textbf{+0.54} \\
DTD          & 60.31 & 61.39 & \textbf{+1.08} \\
EuroSAT      & 64.15 & 74.25 & \textbf{+10.10} \\
\bottomrule
\end{tabular}
\caption{\textbf{Effect of weighted random sampling (WRS) on underperforming datasets.}  
We report the harmonic mean accuracy on 4 vulnerable datasets comparing AAPL and WRS-enhanced AAPL.}
\label{table:wrs_ablation}
\end{table}

\subsubsection{AAPL with weighted random sampling}
\label{exp_subsubsec8}
Building on the identification of ``good'' and ``bad'' augmentations, we quantify how augmentation type affects semantic structure using the average silhouette score, a measure of class separation. Lower scores correlate with poorer zero-shot performance in CoCoOp~\cite{zhou2022conditional}, indicating that some augmentations disrupt semantic consistency.

To mitigate this, we adopt a sampling strategy inspired by active learning~\cite{tamkin2022active, ranganathan2017deep, konyushkova2017learning}, where augmentation types with lower silhouette scores are sampled more frequently. Scores are recomputed each epoch, inverted, and softmax-normalized to emphasize underperforming transformations. This active reweighting targets semantically disruptive transformations and allocates more learning capacity to these cases. This approach is particularly beneficial for datasets like DTD, EuroSAT, StanfordCars, and SUN397, and yields consistent gains in base-to-new generalization (Table~\ref{table:wrs_ablation}), including a notable +10.10\% on EuroSAT.

\begin{table}[t]
\centering
\scriptsize
\renewcommand{\arraystretch}{1.15}
\setlength{\tabcolsep}{4pt}
\begin{tabular}{l ccccccccccc c}
\toprule
\textbf{Setting} &
\rotatebox{90}{ImageNet} &
\rotatebox{90}{Caltech101} &
\rotatebox{90}{OxfordPets} &
\rotatebox{90}{StanfordCars} &
\rotatebox{90}{Flowers102} &
\rotatebox{90}{Food101} &
\rotatebox{90}{FGVCAircraft} &
\rotatebox{90}{SUN397} &
\rotatebox{90}{DTD} &
\rotatebox{90}{EuroSAT} &
\rotatebox{90}{UCF101} &
\rotatebox{90}{Avg.} \\
\midrule
Meta Cons-2   & \textbf{73.45} & \textbf{96.46} & 96.24 & \textbf{72.10} & 79.53 & \textbf{90.93} & 19.24 & \textbf{77.80} & \textbf{63.79} & \textbf{71.64} & \textbf{79.21} & 74.58 \\
Meta Cons-4   & 73.34 & 96.41 & 96.39 & 71.77 & 80.70 & 90.77 & 28.31 & 77.78 & 62.35 & 65.71 & 78.51 & 74.73 \\
Delta Cons-2  & 73.34 & \textbf{96.46} & 96.39 & 71.77 & 80.70 & 90.77 & 28.31 & 77.78 & 62.35 & 65.71 & 78.51 & 74.74 \\
\rowcolor{gray!10}
Delta Cons-4  & 73.09 & 95.87 & \textbf{96.44} & 71.70 & \textbf{82.09} & \textbf{91.10} & \textbf{34.27} & 77.60 & 60.31 & 64.15 & 78.10 & \textbf{74.97} \\
\bottomrule
\end{tabular}
\caption{\textbf{Effect of constraint count in AdTriplet loss.}
Comparison of harmonic mean accuracy for different constraint sizes on the \textit{meta token} and \textit{delta meta token}.}
\label{table:adtriplet_constraints}
\end{table}

\begin{figure}[t]
    \centering
    \includegraphics[width=0.8\linewidth]{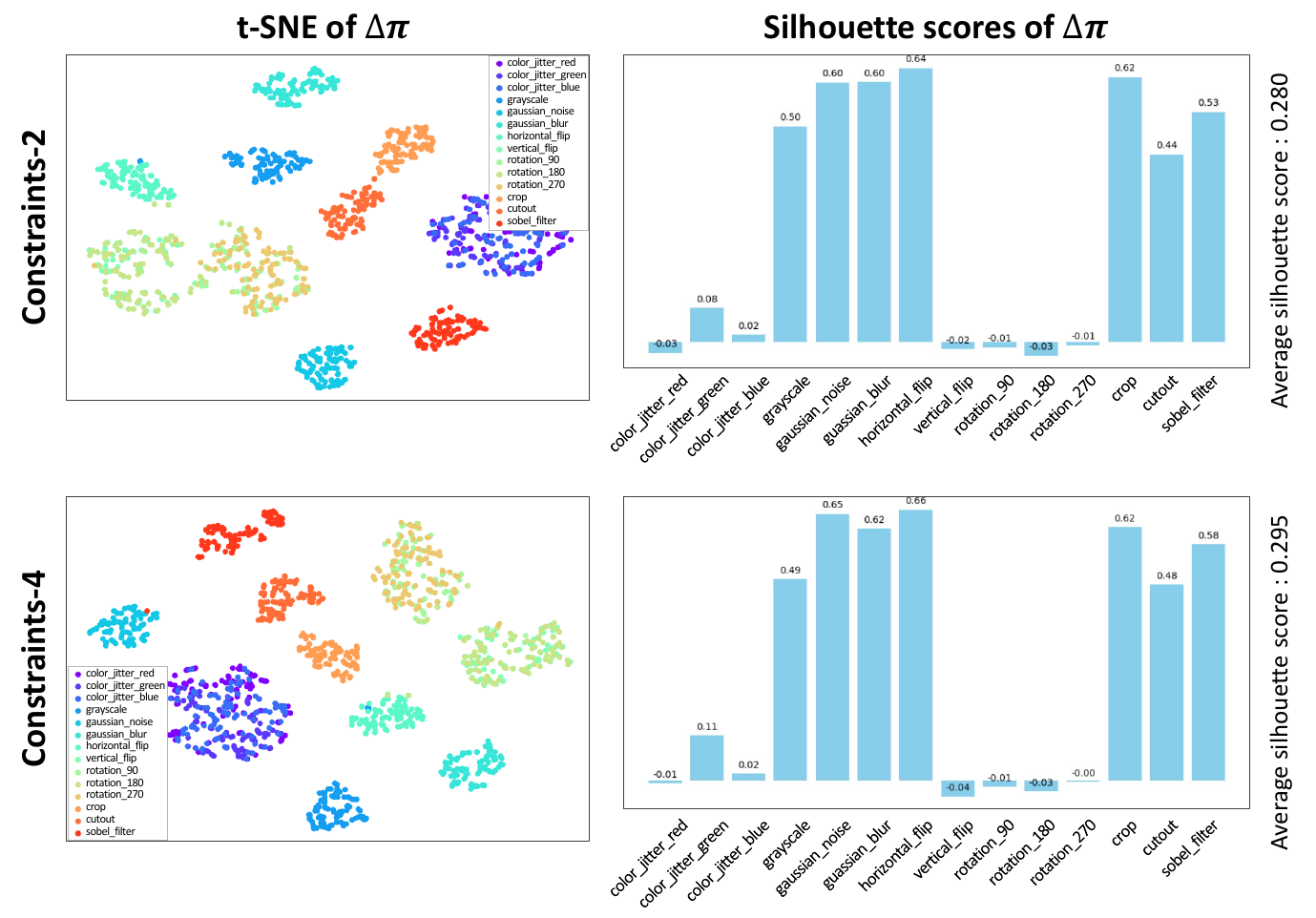}
    \caption{\textbf{The comparison of silhouette score and t-SNE} of the base-to-new generalization for each of the specific augmentation types on FGVCAircraft. All results are from the last epoch.}
    \label{fig:fgvc_constraints}
\end{figure}

\subsection{Ablation on AdTriplet loss constraints}
\label{ablation}
To further examine the role of the AdTriplet loss in attribute-based learning, we conduct an ablation on its constraint settings. Constraints-2 and constraints-4 denote the number of constraints for 4 \textit{delta meta tokens} from 2 different augmentation types, ${\{A, B\}}$, and 2 different classes, ${\{1, 2}\}$, in the AdTriplet loss.The difference between constraints-2 and constraints-4 is illustrated in Fig.~\ref{fig:fig5}. Under constraints-2, following the traditional triplet-loss formulation, a single anchor induces two constraints in the embedding space, while constraints-4 requires two anchors to learn the 4 constraints of \textit{delta meta tokens}. 
As shown in Table~\ref{table:adtriplet_constraints}, constraints-4 performs better than constraints-2. This minimal constraint condition is satisfied by using different attributes and classes as anchors, $\Delta\pi^{1A}$ and $\Delta\pi^{2B}$, which encourages more diverse and separable representations. The equation for the constraints-2 setup is as follows:
\begin{equation}
    L_{AdTriplet}^{const-2} = L_{triplet}(\Delta\pi^{1B}, \Delta\pi^{2B},\Delta\pi^{1A})
    \label{eq:adtriplet_loss_2}     
\end{equation}

In Table~\ref{table:adtriplet_constraints}, for \textit{delta meta token} setup, FGVCAircraft~\cite{maji2013fine}, Flowers102~\cite{nilsback2008automated}, OxfordPets~\cite{parkhi2012cats}, and Food101~\cite{bossard2014food} show better results at constraints-4 compared to constraints-2. In contrast, the remaining data sets perform better in constraints-2. Specifically examining FGVCAircraft within the Delta setup, the silhouette score at constraints-4 is higher than at constraints-2, Fig.~\ref{fig:fgvc_constraints}, indicating a better clustering of augmentation types and improved base-to-new generalization. On average, the silhouette score is 0.280 for constraints-2 and 0.295 for constraints-4, highlighting the clustering advantage of the latter configuration.

\section{Limitation}
AAPL achieves strong performance across diverse benchmarks but depends heavily on the backbone’s ability to encode fine-grained semantics, making it less effective in abstract or visually noisy scenarios. Performance also drops on datasets dominated by broad textures or layout-level structures (\eg DTD and EuroSAT), revealing difficulty in capturing global cues. 
In addition, its effectiveness is influenced by augmentation choice; while well-selected augmentations boost generalization, whereas less informative ones limit gains. Future work includes extending beyond soft prompt tuning to other prompting paradigms, applying AAPL to more complex transformations, and evaluating it on a wider range of vision-language tasks.

\section{Conclusion}
We propose AAPL, a prompt learning framework that disentangles augmentation-specific attributes from class semantics via the \textit{delta meta token} and AdTriplet loss. Augmentation profiling and weighted sampling focus training on challenging transformations, improving generalization in base-to-new, domain-shift, and augmentation robustness settings. Experiments show that AAPL matches or surpasses strong baselines such as CoCoOp, ProGrad, KgCoOp, and DiMPLe while maintaining competitive efficiency. Remaining challenges include handling datasets dominated by global textures or scene layouts and reducing dependence on the choice of augmentations. Addressing these limitations through broader prompting paradigms and evaluations across diverse tasks is a promising avenue for future work.

\section*{Acknowledgments}
This research was financially supported by the Ministry of Trade, Industry and Energy, Korea, under the “Regional Innovation Cluster Development Program(R\&D, p0025331)” supervised by the Korea Institute for Advancement of Technology(KIAT)(50\%). 
This work was supported by the Korea Institute of Energy Technology Evaluation and Planning(KETEP) and the Ministry of Trade, Industry \& Energy(MOTIE) of the Republic of Korea (No. 20224000000100)(40\%). 
This work was supported by the National Research Foundation of Korea(NRF) grant funded by the Korea government(MSIT)(RS-2024-00457860)(10\%). 


\bibliographystyle{elsarticle-num} 
\bibliography{main.bib}

\end{document}